\PassOptionsToPackage{table}{xcolor}
\documentclass[]{selfevolagent}

\usepackage{microtype}
\microtypesetup{expansion=false}
\usepackage{graphicx}
\usepackage{subcaption}
\usepackage{booktabs} 
\usepackage{hyperref}

\usepackage{amsmath}
\usepackage{amssymb}
\usepackage{mathtools}
\usepackage{amsthm}

\usepackage{latexsym}
\usepackage[T1]{fontenc}
\usepackage[utf8]{inputenc}
\usepackage{microtype}
\usepackage{amsmath}
\usepackage{booktabs}
\usepackage{multirow}
\usepackage{amssymb}
\usepackage{balance}
\usepackage{subcaption}

\usepackage{pifont}
\usepackage{makecell}
\usepackage{filecontents}
\usepackage{bbm}
\usepackage{pgfplots}
\usetikzlibrary{patterns}
\usepackage[inline,shortlabels]{enumitem}
\usepackage{natbib}
\usepackage{csquotes}
\usepackage{hyperref}
\usepackage{url}
\usepackage{svg}
\usepackage{bm}
\usepackage{graphics}
\usepackage{graphicx}
\usepackage{array}
\usepackage[english]{babel}
\usepackage{textcomp}
\usepackage{latexsym}
\usepackage{siunitx}  
\usepackage[normalem]{ulem}
\useunder{\uline}{\ul}{}
\usepackage{array}
\usepackage{titletoc}
\usepackage{tikz}
\usepackage{arydshln}
\usepackage[most]{tcolorbox}
\usepackage{algorithm}
\usepackage{algorithmic}
\usepackage{fontawesome5}
\usepackage{wrapfig}
\usetikzlibrary{tikzmark}
\makeatletter
\newcommand*\myfontsize{%
\@setfontsize\myfontsize{7}{8}%
}
\makeatother

\definecolor{geminiBlue}{HTML}{8E8ED7}
\definecolor{qwenBlue}{HTML}{78A2E0}

\definecolor{myred}{rgb}{0.7, 0.3, 0.0}
\definecolor{myblue}{HTML}{0a41b8}
\definecolor{mygreen}{HTML}{056b34}
\definecolor{mypurple}{HTML}{5d1e8b}

\usepackage{graphicx}
\usepackage{booktabs}
\usepackage{booktabs,multirow,array,colortbl,xcolor,pifont}
\usepackage{makecell}
\usepackage{adjustbox}
\usepackage{svg}

\newcommand{\best}[1]{\textbf{\textcolor{black}{#1}}}

\definecolor{dividergray}{RGB}{240,240,240}
\definecolor{ourslavender}{RGB}{239,237,255}   
\definecolor{headergray}{RGB}{250,250,250}
\newcommand{\qwenicon}{%
  \raisebox{-0.2\height}{\includesvg[height=1.15em]{figures/Qwen3}}\hspace{0.25em}%
}
\newcommand{\gpticon}{%
  \raisebox{-0.2\height}{\includesvg[height=1.15em]{figures/openai}}\hspace{0.25em}%
}

\usepackage{listings}
\usepackage{xcolor}

\definecolor{codegreen}{rgb}{0,0.6,0}
\definecolor{codegray}{rgb}{0.5,0.5,0.5}
\definecolor{codepurple}{rgb}{0.58,0,0.82}
\definecolor{backcolour}{rgb}{0.95,0.95,0.92}

\lstdefinestyle{pystyle}{
    backgroundcolor=\color{backcolour},   
    commentstyle=\color{codegreen},
    keywordstyle=\color{magenta},
    numberstyle=\tiny\color{codegray},
    stringstyle=\color{codepurple},
    basicstyle=\ttfamily\small,
    breakatwhitespace=false,         
    breaklines=true,                 
    captionpos=b,                    
    keepspaces=true,                 
    numbers=left,                    
    numbersep=5pt,                  
    showspaces=false,                
    showstringspaces=false,
    showtabs=false,                  
    tabsize=4
}
\definecolor{myblue}{RGB}{108, 142, 191}

\usepackage{fontawesome5}
\usepackage{enumitem}

\theoremstyle{plain}
\newtheorem{theorem}{Theorem}[section]

\theoremstyle{definition}
\newtheorem{definition}[theorem]{Definition}

\theoremstyle{remark}

\usepackage[textsize=tiny]{todonotes}
\usepackage{xcolor}

\definecolor{startBlue}{HTML}{1628a7}  
\definecolor{endPurple}{HTML}{8b16aa}
\newcommand{\elegantMuSE}{%
\textcolor{startBlue}{\textbf{M}}
\textcolor{startBlue!60!endPurple}{\textbf{u}}
\textcolor{endPurple!60!startBlue}{\textbf{S}}
\textcolor{endPurple}{\textbf{E}}
}

\definecolor{wx}{RGB}{54, 89, 170}

\title{\elegantMuSE Agent: A Multimodal Reasoning Agent with Stateful Experiences}
\author[1,2,3,\ddagger,*]{Shijian Wang}
\author[3,\ddagger]{~Jiarui Jin}
\author[2]{~Runhao Fu}
\author[3,4,*]{~Zexuan Yan}
\author[2]{~Xingjian Wang}
\author[3,5,*]{~Mengkang Hu} 
\author[3,6,*]{\\Eric Wang}
\author[3,7,*]{~Xiaoxi Li}
\author[3,4,*]{~Kangning Zhang}
\author[1]{~Li Yao}
\author[3\dagger]{~Wenxiang Jiao}
\author[2\dagger]{~Xuelian Cheng}
\author[3\dagger]{\\Yuan Lu}
\author[2]{~Zongyuan Ge}

\affiliation[1]{Southeast University}
\affiliation[2]{Monash University}
\affiliation[3]{Xiaohongshu Inc.}
\affiliation[4]{Shanghai Jiao Tong University}
\affiliation[5]{University of Hong Kong}
\affiliation[6]{Zhejiang University}
\affiliation[7]{Renmin University of China}

\contribution[*]{Work done during internship at Xiaohongshu Inc.}
\contribution[\ddagger]{Equal contribution}
\contribution[\dagger]{Corresponding authors}

\metadata[\raisebox{-0.18ex}{\scalebox{0.89}{\faEnvelope}}~Contact]{shijian@seu.edu.cn}
\metadata[{\faGithub}~Code]{\href{https://github.com/DeepExperience/MuSEAgent}{https://github.com/DeepExperience/MuSEAgent}} 

\newtcolorbox[auto counter, number within=section]{promptbox}[2][]{%
  colback=white, 
  colframe=myblue,  
  width=\textwidth,
  arc=2mm, 
  title={\normalsize\faInfoCircle\hspace{0.5em}#2},
  breakable,
  fonttitle=\bfseries\Large, 
  fontupper=\small, 
  drop shadow southeast, 
  top=2mm,
  bottom=2mm,
  before skip=3mm,
  after skip=3mm,
  boxrule=0.5mm,
  #1
}

\abstract{Research agents have recently achieved significant progress in information seeking and synthesis across heterogeneous textual and visual sources. In this paper, we introduce \textbf{MuSEAgent}, a multimodal reasoning agent that enhances decision-making by extending the capabilities of research agents to discover and leverage stateful experiences. Rather than relying on trajectory-level retrieval, we propose a stateful experience learning paradigm that abstracts interaction data into atomic decision experiences through hindsight reasoning. These experiences are organized into a quality-filtered experience bank that supports policy-driven experience retrieval at inference time. Specifically, MuSEAgent enables adaptive experience exploitation through complementary wide- and deep-search strategies, allowing the agent to dynamically retrieve multimodal guidance across diverse compositional semantic viewpoints. Extensive experiments demonstrate that MuSEAgent consistently outperforms strong trajectory-level experience retrieval baselines on both fine-grained visual perception and complex multimodal reasoning tasks. These results validate the effectiveness of stateful experience modeling in improving multimodal agent reasoning.}

\begin{document}
\maketitle

\section{Introduction}
\label{sec:intro}
Multimodal agents that integrate vision and language have achieved significant progress in perception and generation across heterogeneous modalities \citep{he2024webvoyager, xie2024osworld, yang2023mm}. A core challenge facing these agents is learning to effectively exploit tools in order to navigate complex, multi-step environments. 
One promising direction in large language model research is to retrieve and learn from similar past experiences stored in a memory bank \citep{zhou2025memento, wang2025memento}. 
Recent work further explores experience-augmented agents that leverage historical interactions to improve autonomy and generalization \citep{zhao2024expel, shinn2023reflexion, wang2023voyager}. 
However, extending these ideas to multimodal agents introduces two fundamental challenges. 
First, visual inputs typically carry much lower information density than textual ones, such that retrieving entire interaction histories often injects redundant or irrelevant context, amplifying reasoning noise under constrained context windows \citep{liu2024lost, chang2024agentboard}. 
Second, multimodal reasoning requires interleaved thinking across diverse modalities, which makes identifying relevant and similar memory cases substantially more difficult. 
Moreover, trajectory-level experience retrieval fails to provide fine-grained decision guidance when agents encounter intermediate reasoning bottlenecks, where state-specific tactical knowledge is far more beneficial than coarse task-level analogies \citep{zhang2025appagent, xie2024osworld}.

In this paper, we propose MuSEAgent, a novel \textbf{Mu}ltimodal Reasoning Agent with \textbf{S}tateful \textbf{E}xperiences.
Specifically, we reformulate multimodal agent reasoning as a state-aware experience learning process, in which historical trajectories are abstracted into atomic state-action pairs through hindsight reasoning. 
These abstracted experiences are organized into a quality-filtered experience bank that supports retrieval conditioned on the agent's current decision state, rather than on static initial task contexts, enabling more precise and noise-free guidance at each reasoning step.

To effectively exploit stateful experiences during reasoning, MuSEAgent extends the capabilities of deep research agents to iteratively query the experience bank during inference.
Specifically, we design a compositional state representation mechanism that decomposes complex multimodal states into multiple semantic viewpoints, enabling experience indexing across perceptual intent, tool execution history, and interaction context. 
Based on this representation, the agent performs \textit{Wide Search} to retrieve cross-task strategic knowledge and \textit{Deep Search} to iteratively refine retrieval across complementary semantic viewpoints within a single reasoning step.

Experimental results demonstrate that MuSEAgent consistently outperforms strong trajectory-level experience retrieval baselines by nearly 8\% in average accuracy, particularly on fine-grained multimodal reasoning tasks where state-level guidance effectively mitigates contextual noise.

Our contributions are summarized as follows:
\begin{itemize}
\item We propose a stateful experience learning framework for multimodal agents by abstracting high-quality atomic decision experiences from historical trajectories via hindsight reasoning.
\item We design a novel Deep-and-Wide experience search mechanism that enables adaptive retrieval across compositional semantic viewpoints.
\item We demonstrate that MuSEAgent achieves significant improvements on fine-grained perception and complex multimodal reasoning tasks compared with trajectory-based agents.
\end{itemize}

\section{Related Work}

\paragraph{Visual Reasoning.}
Multimodal Large Language Models (MLLMs), such as GPT-4o\citep{hurst2024gpt}, LLaVA series\citep{liu2023visual, liu2024improved, liu2024llavanext} and Qwen-VL series\citep{bai2023qwen, Qwen3-VL, qwen3.5}, have advanced multimodal understanding through large-scale vision-language pre-training. However, they remain fragile in multi-step visual reasoning. Evaluations on benchmarks such as V* Bench\citep{wu2024v}, HR-Bench\citep{wang2025divide}, MME-RealWorld-Lite\citep{zhang2024mme}, and ZoomBench\citep{wei2026zooming} reveal that these models still exhibit persistent logical inconsistencies and hallucinations when detailed visual grounding is required. To improve visual reasoning, approaches such as LLaVA-CoT\citep{xu2025llava} and LlamaV-o1\citep{thawakar2025llamav} introduce Chain-of-Thought prompting to decompose problems into sequential steps. Nevertheless, as analyzed in Insight-V\citep{dong2025insight}, current models struggle to maintain consistent intermediate representations over long reasoning chains, often leading to contradiction or collapse. To address these limitations, our MuSEAgent models visual reasoning as an iterative state-based process, learning from fine-grained state-level experiences to achieve structured refinement.

\paragraph{Multimodal Agent.}
Recent advances in multimodal agents shift visual reasoning from single-pass inference to interactive decision-making with tool use and planning\citep{xi2025rise, wang2024survey}. Inspired by ReAct\citep{yao2022react}, prior works enable LLMs to coordinate vision modules or structured programs for complex visual reasoning\citep{yang2023mm, gupta2023visual, suris2023vipergpt, shen2023hugginggpt}. While improving modularity, these systems typically append entire interaction histories or follow rigid execution trajectories\citep{he2024webvoyager, zhang2025appagent}, which introduces contextual redundancy and undermines long-horizon consistency. To address this limitation, our MuSEAgent formalizes agentic multimodal reasoning as a Markov Decision Process over discrete state units, converting interaction histories into fine-grained experiences.

\paragraph{Experience-driven Agent Learning.}
Recent work explores memory-augmented agents to improve long-horizon reasoning by reusing past trajectories\citep{shinn2023reflexion, wang2023voyager, zhao2024expel, packer2023memgpt}. However, most existing methods retrieve past experiences at the coarse, trajectory level\citep{yao2023retroformer, zhu2023ghost}. Because entire trajectories are long and rigid, directly matching them to a new problem often introduces irrelevant noise, making it difficult for agents to flexibly adapt these experiences to fine-grained multimodal reasoning\citep{yao2022react, yang2023mm}. Our MuSEAgent addresses this limitation by modeling multimodal reasoning as a Markov Decision Process over Stateful Experiences. Instead of relying on full trajectories, MuSEAgent abstracts them into discrete, reusable state units, enabling fine-grained, state-level experience retrieval for long-horizon visual reasoning.

\section{MuSEAgent}
\label{sec:method}

\begin{figure}[t]
    \centering    \includegraphics[width=0.9\linewidth]{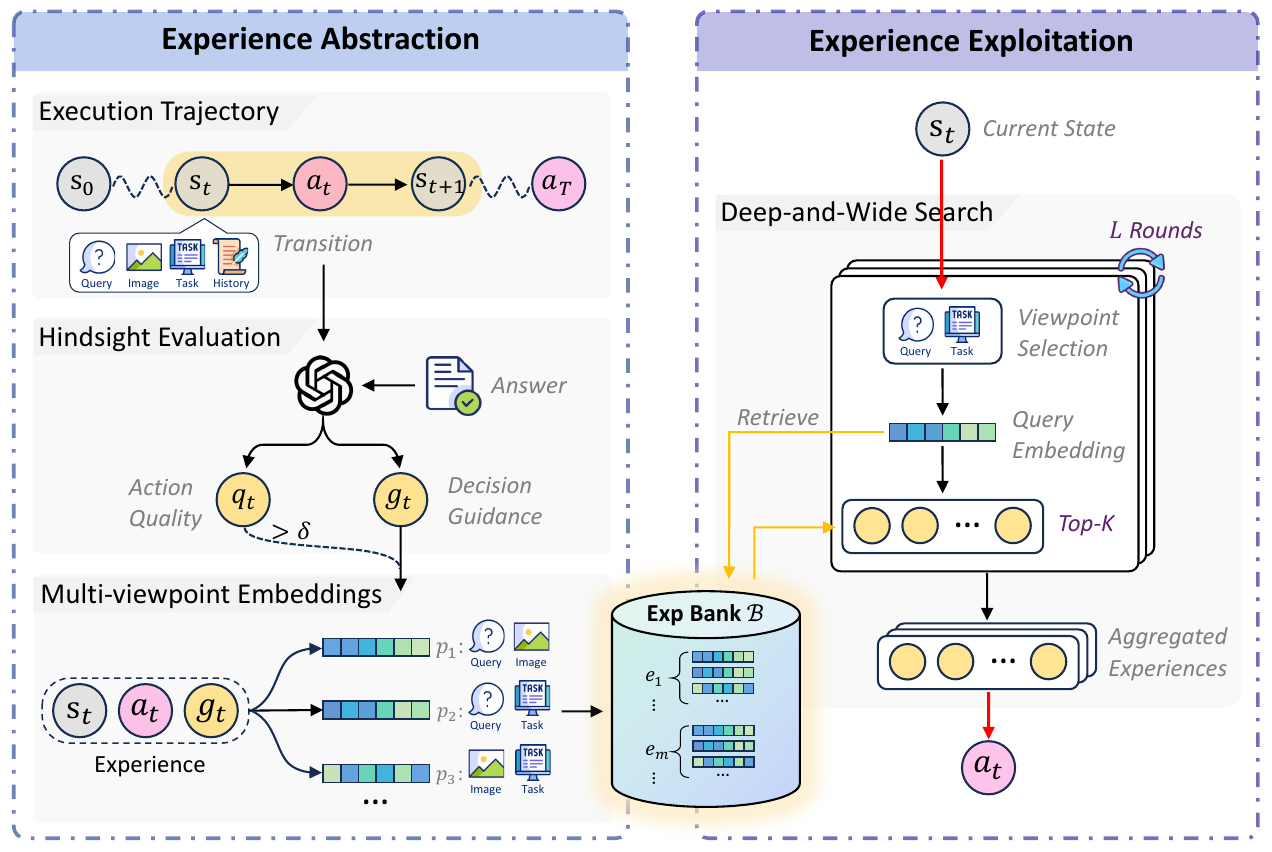}
    \caption{The overview of MuSEAgent. The framework consists of two phases: (1) \textbf{Experience Abstraction}, which extracts state-level experiences via hindsight evaluation and builds multi-viewpoint embeddings for each experience; (2) \textbf{Experience Exploitation}, where the agent performs a deep-and-wide search over the experience bank to determine the next action. }
    \label{fig:workflow}
\end{figure}

We present MuSEAgent, illustrated in Figure~\ref{fig:workflow}, a novel framework designed to enhance the reasoning capabilities of multimodal agents through the abstraction and exploitation of stateful experiences.
Specifically, we formulate the agentic decision-making process as a stateful Markov Decision Process (MDP) in Section~\ref{sec:problem_formulation}.
In Section~\ref{sec:abstraction}, we introduce a hindsight reasoning mechanism that abstracts high-quality, state-aware experiences from historical interactions, mitigating the redundancy and noise commonly present in conventional trajectory-level experiences.
Building upon this abstraction, we introduce a compositional state representation that generates multiple embeddings for each experience using different combinations of multimodal state representations (textual questions and task instructions, visual inputs, and structured action sequences). As a result, a single experience can be retrieved through multiple state-aware viewpoints, improving retrieval flexibility and coverage.
Finally, Section~\ref{sec:deepsearch} details an online experience exploration process that performs deep-and-wide search over the experience bank, enabling the agent to iteratively retrieve relevant experiences and synthesize fine-grained guidance for decision-making.

\subsection{Problem Formulation}
\label{sec:problem_formulation}

In this paper, we integrate LLM agents with multimodal experience-based reasoning, a paradigm in which new problems are solved by leveraging solutions to previously encountered, similar problems stored in the experience bank.  
To formally characterize the sequential decision-making process of MuSEAgent, we model agentic reasoning as a stateful experience-driven decision process.

\begin{definition}[Stateful Experience-Based MDP]
Let $u \in \mathcal{U}$ denote the natural language query provided at the beginning. 
The agent interacts with multimodal observations and historical execution contexts across multiple reasoning steps. 
The decision state at step $t$ is defined as $s_t = (u, v_t, d, H_t)$,
where $u$ is the fixed user instruction, $v_t$ denotes the current visual observation or perceptual input, $d$ is an optional task descriptor, and $H_t$ is the execution history up to step $t$.
The action space is defined as
$a_t \in \mathcal{A}_{\text{tool}} \cup \mathcal{A}_{\text{exp}}$,
where $\mathcal{A}_{\text{tool}}$ contains task-specific execution tools and $\mathcal{A}_{\text{exp}}$ corresponds to experience retrieval over an experience bank.
A trajectory generated by the agent is represented as
$\tau = (s_1,a_1,s_2,a_2,\dots,s_T,a_T)$,
where $T$ denotes the task-dependent reasoning horizon. 
Instead of directly using raw trajectories as experiences, which often contain redundant multimodal signals, we aim to abstract high-quality decision knowledge into reusable, state-level experiences that can be retrieved and leveraged across tasks and reasoning steps.
\end{definition}

\subsection{Stateful Experience Abstraction via Hindsight Reasoning}
\label{sec:abstraction}

\paragraph{Experience Abstraction.}
To construct a compact and noise-resistant experience bank, we first decompose agent trajectories into atomic state–action transitions $(s_t, a_t, s_{t+1})$. These transitions serve as candidate experience units that capture localized decision contexts within long multimodal interaction trajectories.
Each transition is then evaluated using a hindsight reasoning model that assesses the quality of the decision and extracts reusable decision guidance. Concretely, a multimodal reasoning model $\mathcal{Q}\phi$ (e.g., GPT-4o) takes the transition context as input and produces both a scalar quality score $q_t$ and a textual guidance summary $g_t$:
\begin{equation}
(q_t, g_t) = \mathcal{Q}_\phi(s_t, a_t, s_{t+1}),
\end{equation}
where $q_t \in [0,10]$ reflects the estimated decision quality of the action taken at state $s_t$, and $g_t$ summarizes the key decision experience abstracted from this transition.

To suppress noisy or uninformative transitions, we retain only high-quality evaluated transitions to construct the experience bank:
\begin{equation}
\mathcal{B} = \{e_t = (s_t, a_t, g_t) \mid q_t \ge \delta\},
\end{equation}
where $\delta$ is a predefined quality threshold (default $\delta = 5.0$). Each retained transition is thus converted into a reusable experience unit consisting of the decision context $(s_t)$, the executed action $(a_t)$, and the abstracted decision guidance $(g_t)$.
During offline indexing, the state component $s_t$ of each experience $e_t$ is further decomposed into multiple semantic viewpoints for embedding. Consequently, subsequent retrieval operates over these state embeddings to find relevant experiences, while the returned guidance $g_t$ provides the actionable decision insight.

\paragraph{Compositional State Representation.}
Multimodal agent states are heterogeneous, including textual instructions, visual observations, task descriptors, and execution histories, which provide complementary signals for experience retrieval.
To enable flexible querying over such heterogeneous states, we organize each state into multiple semantic viewpoints. Let $\mathcal{P} = \{p_1, p_2, \dots, p_M\}$ denote a set of predefined viewpoints, where each viewpoint corresponds to a specific composition of state components.
During the offline experience indexing stage, each experience $e_t$ is associated with multiple embeddings derived from the corresponding state components under each viewpoint:
\begin{equation}
\mathbf{z}_t^{(i)} = f_\theta(p_i(s_t)), \quad p_i \in \mathcal{P},
\end{equation}
where $f_\theta$ denotes a multimodal embedding model and $p_i(s_t)$ extracts the state elements for viewpoint $p_i$.
As a result, each experience $e_t$ is indexed under multiple complementary semantic perspectives, enabling the agent to retrieve relevant experiences through different contextual cues. At inference time, the agent policy adaptively selects an appropriate viewpoint to query the experience bank.

Algorithm~\ref{alg:abstraction} summarizes the complete experience abstraction procedure.

\begin{algorithm}[tb]
\caption{Stateful Experience Abstraction via Hindsight Reasoning}
\label{alg:abstraction}
\begin{algorithmic}[1]
\REQUIRE Historical trajectories $\mathcal{T} = \{\tau_1, \tau_2, \dots\}$, 
         quality threshold $\delta$, 
         candidate semantic viewpoints $\mathcal{P} = \{p_1, \dots, p_M\}$.
\ENSURE  Updated Experience Bank $\mathcal{B}$.

\STATE \textcolor{myblue}{\# \textbf{Filter high-quality transitions and index each experience under $M$ semantic viewpoints.}}
\STATE Initialize experience bank $\mathcal{B} \leftarrow \emptyset$
\FOR{each trajectory $\tau \in \mathcal{T}$}
    \FOR{each atomic transition $(s_t, a_t, s_{t+1}) \in \tau$}
        \STATE Compute hindsight quality score and guidance summary:
               $(q_t, g_t) \leftarrow \mathcal{Q}_\phi(s_t, a_t, s_{t+1})$
        \IF{$q_t \ge \delta$}
            \STATE Construct stateful experience $e_t \leftarrow (s_t, a_t, g_t)$
            \STATE Build multi-viewpoint embeddings:
                   $\mathbf{z}_t^{(i)} \leftarrow f_\theta(p_i(s_t)),\ \forall\, p_i \in \mathcal{P}$
            \STATE Update experience bank $\mathcal{B} \leftarrow \mathcal{B} \cup \{e_t\}$
        \ENDIF
    \ENDFOR
\ENDFOR
\RETURN $\mathcal{B}$
\end{algorithmic}
\end{algorithm}

\begin{algorithm}[tb]
\caption{Experience Exploitation via Deep-and-Wide Search}
\label{alg:exploitation}
\begin{algorithmic}[1]
\REQUIRE Experience Bank $\mathcal{B}$, current state $s_t$,
         retrieval breadth $K$, max refinement rounds $L$,
         candidate semantic viewpoints $\mathcal{P} = \{p_1, \dots, p_M\}$.
\ENSURE  Execution action $a_t$.

\STATE Initialize unified guidance set $\mathcal{E}_{\text{deep-wide}} \leftarrow \emptyset$

\STATE \textcolor{myblue}{\# \textbf{Deep Search: Iterative Semantic Viewpoint Refinement}}
\FOR{round $j = 1$ to $L$}
    \STATE Sample semantic viewpoint and formulate query embedding:
           $p_{i_j} \sim \pi(\cdot \mid s_t)$,\quad
           $\mathbf{q}_t^{(i_j)} \leftarrow f_\theta\!\left(p_{i_j}(s_t)\right)$
    \STATE \textcolor{myblue}{\# \textbf{Wide Search: Top-$K$ Retrieval under Current Viewpoint}}
    \STATE Retrieve top-$K$ experiences:
    \[
        \mathcal{E}^{(j)} \leftarrow 
        \operatorname{Top\text{-}K}_{e_m \in \mathcal{B}}
        \Bigl(\operatorname{sim}\!\left(\mathbf{q}_t^{(i_j)},\, 
        \mathbf{z}_m^{(i_j)}\right)\Bigr)
    \]
    \STATE Aggregate retrieved experiences:
           $\mathcal{E}_{\text{deep-wide}} \leftarrow 
            \mathcal{E}_{\text{deep-wide}} \cup \mathcal{E}^{(j)}$
\ENDFOR

\STATE \textcolor{myblue}{\# \textbf{Action Generation}}
\STATE Generate execution action:
       $a_t \sim \pi\!\left(\cdot \mid s_t,\, \mathcal{E}_{\text{deep-wide}}\right)$
\RETURN $a_t$
\end{algorithmic}
\end{algorithm}

\subsection{Stateful Experience Exploitation via Deep-and-Wide Search}
\label{sec:deepsearch}

At inference step $t$, before committing to a tool-specific action, the agent consults the experience bank to obtain additional decision guidance. 
The agent policy $\pi(\cdot \mid s_t)$ first selects a semantic viewpoint $p_i \in \mathcal{P}$ based on the current state $s_t$.
Since each experience $e_i$ in the experience bank $\mathcal{B}$ is indexed with viewpoint-specific embeddings, the selected viewpoint determines which embedding index is queried. 
The agent then constructs a query embedding and retrieves relevant experiences from the corresponding viewpoint index.
%
To balance generalization and precise state matching, we adopt two complementary retrieval strategies: \textit{Wide~Search}, and \textit{Deep~Search}.

\paragraph{Wide Search.}
To identify broadly relevant experiences, Wide~Search performs breadth-oriented retrieval under the currently selected viewpoint. 
Given the query embedding $\mathbf{q}_t^{(i)} = f_\theta(p_i(s_t))$ constructed under the selected viewpoint $p_i$, the agent retrieves the Top-$K$ most relevant experiences 
from the corresponding viewpoint index according to cosine similarity:
\begin{equation}
\mathcal{E}_{\text{wide}}(s_t)
=
\operatorname{Top\text{-}K}_{e_m \in \mathcal{B}}
\;
\operatorname{sim}
\left(
\mathbf{q}_t^{(i)}, \mathbf{z}_m^{(i)}
\right),
\end{equation}
where $\mathbf{z}_m^{i}$ denotes the stored embedding of experience $e_m$ under the selected viewpoint, and $K$ controls the retrieval breadth. 
By retrieving multiple experiences with similar contextual signals, Wide~Search exposes the agent to diverse decision contexts and helps identify reusable reasoning patterns that generalize across related tasks.

\paragraph{Deep Search.}
When broader retrieval does not provide sufficiently precise guidance, Deep~Search performs iterative refinement by querying the experience bank under multiple semantic viewpoints.
At each refinement round $j$, the agent selects a viewpoint 
$p_{i_j} \in \mathcal{P}$ and constructs a viewpoint-specific query embedding
\begin{equation}
\mathbf{q}_t^{(i_j)} = f_\theta(p_{i_j}(s_t)).
\end{equation}
The agent then retrieves the most relevant experience under that viewpoint:
\begin{equation}
\mathcal{E}_{\text{deep}}(s_t)
=
\bigcup_{j=1}^{L}
\operatorname{Top\text{-}1}_{e_m \in \mathcal{B}}
\;
\operatorname{sim}
\left(
\mathbf{q}_t^{(i_j)}, \mathbf{z}_m^{(i_j)}
\right),
\end{equation}
where $L$ denotes the maximum number of refinement rounds.
Each round emphasizes a different aspect of the state representation, allowing the agent to progressively align task intent, perceptual observations, and historical tool usage patterns. For example, the agent may initiate retrieval under a purely visual viewpoint to resolve bounding-box ambiguities, and subsequently issue a secondary query under the execution histories (i.e., tool) viewpoint to validate tool parameter syntax.

\paragraph{Unified Deep-and-Wide Search.}
In practice, experience search is performed iteratively under a sequence of viewpoints selected by the agent. At round $j$, a viewpoint $p_{i_j}$
is chosen, and the agent retrieves the Top-$K$ most relevant
experiences under that viewpoint. The final experience set is
\begin{equation}
\mathcal{E}_{\text{deep-wide}}(s_t)
=
\bigcup_{j=1}^{L}
\operatorname{Top\text{-}K}_{e_m \in \mathcal{B}}
\;
\operatorname{sim}
\left(
\mathbf{q}_t^{(i_j)}, \mathbf{z}_m^{(i_j)}
\right),
\end{equation}
where $L$ is the number of retrieval rounds and $K$ controls the
retrieval breadth per round. The retrieved experience strings $\{g_m\}$ are injected into the agent's working context to decide the next action $a_t \sim \pi(\cdot \mid s_t, \mathcal{E}_{\text{deep-wide}})$.

Algorithm~\ref{alg:exploitation} summarizes the complete online experience exploitation procedure.

\section{Experiments}

To demonstrate the efficacy of MuSEAgent, we design our empirical investigation around four primary aspects:
\begin{enumerate}
    \item \textbf{Overall Performance:} \textit{Can stateful experience search outperform trajectory-level baselines in multimodal reasoning tasks?} We compare MuSEAgent against established reasoning and trajectory-based methods across diverse benchmarks~(Sec.~\ref{sec:performance}).
    \item \textbf{Deep-and-Wide Search:} \textit{How does the Deep-and-Wide search mechanism contribute to agent performance?} We analyze the impact of scaling the depth and breadth of experience search on agent performance~(Sec.~\ref{sec:deepsearch}).
    \item \textbf{Generalization:} \textit{Do stateful experiences generalize to out-of-distribution domains?} We evaluate the transferability of learned stateful experience to unseen domain~(Sec.~\ref{sec:ood}).
    \item \textbf{Ablation:} \textit{What is the effect of key hyperparameters and component choices?} We conduct ablation studies on experience sources, hindsight models, and quality score thresholds~(Sec.~\ref{sec:ablation}).
\end{enumerate}

\subsection{Experimental Setup}

\begin{table}[t] 
\caption{Statistics of Benchmark Datasets.}
\label{tab:data}
\centering
\setlength{\tabcolsep}{3pt}

\begin{adjustbox}{max width=\linewidth} 
\begin{tabular}{c ccc ccc cc}
\toprule
\multirow{2}{*}{\textbf{Split}} &
\multicolumn{3}{c}{\textbf{V* Bench}} &
\multicolumn{3}{c}{\textbf{MME-RealWorld-Lite}} &
\multirow{2}{*}{\textbf{Zoom-Bench}} &
\multirow{2}{*}{\textbf{HR-Bench}} \\
\cmidrule(lr){2-4}\cmidrule(lr){5-7}
& \makecell{\textbf{Direct Attr.}} & \makecell{\textbf{Rel. Pos.}} & \textbf{All}
& \textbf{Percept.} & \textbf{Reas.} & \textbf{All}
& & \\
\midrule
Exploration & 57 & 38 & 95 & 584 & 375 & 959 & 310 & 100 \\
Evaluation  & 58 & 38 & 96 & 585 & 375 & 960 & 311 & 100 \\
\bottomrule
\end{tabular}
\end{adjustbox}
\end{table}

\paragraph{Benchmark Datasets.}
To evaluate multimodal reasoning capabilities, we test on four multiple-choice VQA benchmarks: V*~Bench ~\citep{wu2024v}, MME-RealWorld-Lite ~\citep{zhang2024mme}, ZoomBench ~\citep{wei2026zooming} and HR-Bench ~\citep{wang2025divide}, spanning diverse domains including fine-grained visual perception, real-world visual understanding, and complex multimodal reasoning. We partition each dataset into a 1:1 exploration and evaluation split, with details presented in Table~\ref{tab:data}. The agent interacts with the exploration split to construct the experience bank, and we report the accuracy on the evaluation split to assess the final agent performance. Detailed dataset descriptions are provided in Appendix~\ref{app:dataset}.

\paragraph{Baseline Methods.}
To benchmark the effectiveness of stateful experiences, we compare MuSEAgent against four representative baselines, encompassing vanilla reasoning, Tool-Integrated Reasoning (TIR)~\citep{lin2025understanding}, and trajectory-level experience-based methods. These include Vanilla CoT ~\citep{kojima2022large} operating without external tools, ReAct ~\citep{yao2022react} incorporating dynamic tool usage, Reflexion ~\citep{shinn2023reflexion}, which derives reflective experiences from failed trajectories, and Expel ~\citep{zhao2024expel}, which extracts insights from both successful and unsuccessful trajectories. Detailed descriptions of these baseline methods are presented in Appendix~\ref{app:baselines}.

\paragraph{Tool Bank.}
We equip the agent with a comprehensive suite of 13 multimodal tools tailored for fine-grained perception and external knowledge acquisition. These encompass broad functional categories, ranging from basic extraction, computation, and retrieval (OCR, mathematical equation solving, standard calculations, web search) to advanced visual processing (object localization, image zoom-in, image cropping, visual region highlighting, region depth estimation, object depth estimation) and cross-modal semantic alignment (image-to-image similarity, image-to-text similarity, and text-to-image similarity). Complete tool specifications are available in the Appendix~\ref{app:tool}.

\paragraph{Implementation Details.}
To ensure strong instruction following and tool use, we employ three advanced agentic MLLMs as base models: Qwen3-VL-32B-Instruct, Qwen3-VL-235B-A22B-Instruct~\citep{Qwen3-VL}, and Qwen3.5-397B-A17B~\citep{qwen3.5}. Unless otherwise stated, we employ GPT-4o~\citep{hurst2024gpt} as the hindsight reasoning model to abstract stateful experiences from both correct and incorrect trajectories. We utilize Qwen3-VL-8B-Embedding~\citep{li2026qwen3} to encode multimodal states into a unified vector space. During the online reasoning phase, we configure the Deep-and-Wide search mechanism with a maximum iteration depth of 3 across varying semantic viewpoints, retrieving exactly 3 distinct experiences per search. We establish a quality score threshold of 5.0 to filter suboptimal historical experiences. The specific effects of these configurations, namely the experience source, hindsight model choice, and quality score threshold, are systematically analyzed in Sec.~\ref{sec:ablation}.
All prompts are summarized in Appendix~\ref{app:prompt}.

\begin{table*}[t]
\caption{Performance comparison of MuSEAgent against baselines on various benchmarks under three base models. We report the accuracy (\%) for all tasks. Best results are highlighted in \best{bold}.}
\centering
\fontsize{8pt}{10pt}\selectfont
\setlength{\tabcolsep}{3pt}

\begin{adjustbox}{max width=\textwidth}
\begin{tabular}{
  >{\raggedright\arraybackslash}l
  c
  c c c
  c c c
  c c c
}
\toprule
\multirow{2}{*}{\textbf{Method}} &
\multirow{2}{*}{\textbf{Exp. Level}} &
\multicolumn{3}{c}{\textbf{V* Bench}} &
\multicolumn{3}{c}{\textbf{MME-RealWorld-Lite}} &
\multirow{2}{*}{\textbf{Zoom-Bench}} &
\multirow{2}{*}{\textbf{HR-Bench}} &
\multirow{2}{*}{\textbf{Avg.}} \\
\cmidrule(lr){3-5}\cmidrule(lr){6-8}
&
&
\textbf{Direct~Attr.} & \textbf{Rel. Pos.} & \textbf{Micro Avg.} &
\textbf{Percep.} & \textbf{Reas.} & \textbf{Micro Avg.} &
& & \\
\midrule

\rowcolor{dividergray}
\multicolumn{11}{c}{\qwenicon \textbf{\textit{Qwen3-VL-32B-Instruct}}} \\

Vanilla CoT~\citep{kojima2022large}
  & N/A 
  & 27.59  & 57.89  & 39.58 
  & 28.55  & 28.53  & 28.54 
  & 31.83  & 51.00  & 37.74  \\

w/ ReAct~\citep{yao2022react}
  & N/A
  & 62.07  & 84.21  & 70.83 
  & 44.44  & 36.80  & 41.46 
  & 44.69  & 66.00  & 55.75  \\

w/ Reflexion~\citep{shinn2023reflexion}
  & Trajectory
  & {74.13} & {76.32} & {75.00}
  & {46.15} & {34.13} & {41.46}
  & {48.87} & {64.00} & {57.33} \\

w/ Expel~\citep{zhao2024expel}
  & Trajectory
  & 70.69  & 78.95  & 73.96 
  & 42.39  & \best{40.27} & 41.56 
  & 46.30  & 66.00  & 56.96  \\

\rowcolor{ourslavender}
w/ MuSEAgent
  & State
  & \best{86.21} & \best{84.21} & \best{85.42}
  & \best{46.32} & 38.13  & \best{43.13}
  & \best{54.66} & \best{78.00} & \best{65.30} \\
\midrule

\rowcolor{dividergray}
\multicolumn{11}{c}{\qwenicon \textbf{\textit{Qwen3-VL-235B-A22B-Instruct}}} \\

Vanilla CoT~\citep{kojima2022large}
  & N/A
  & 24.14  & 47.37  & 33.33 
  & 33.68  & 29.87  & 32.19 
  & 36.98  & 54.00  & 39.13  \\

w/ ReAct~\citep{yao2022react}
  & N/A
  & 44.83  & 68.42  & 54.17 
  & 42.74  & 35.20  & 39.79 
  & 43.73  & 63.00  & 50.17  \\

w/ Reflexion~\citep{shinn2023reflexion}
  & Trajectory
  & 48.28  & 86.84  & 63.54 
  & 43.42  & 31.73  & 38.85 
  & 42.44  & 69.00  & 53.46  \\

w/ Expel~\citep{zhao2024expel}
  & Trajectory
  & 53.45  & 73.68  & 61.46 
  & 42.39  & 34.67  & 39.38 
  & 43.09  & 72.00  & 53.98  \\

\rowcolor{ourslavender}
w/ MuSEAgent
  & State
  & \best{55.17} & \best{92.11} & \best{69.79}
  & \best{42.91} & \best{35.73} & \best{40.10}
  & \best{45.02} & \best{74.00} & \best{57.23} \\
\midrule

\rowcolor{dividergray}
\multicolumn{11}{c}{\qwenicon \textbf{\textit{Qwen3.5-397B-A17B}}} \\

Vanilla CoT~\citep{kojima2022large}
  & N/A
  & 39.66  & 60.53  & 47.92 
  & 39.32  & 36.80  & 38.33 
  & 39.23  & 61.00  & 46.62  \\

w/ ReAct~\citep{yao2022react}
  & N/A
  & 79.31  & 71.05  & 76.04 
  & 50.26  & 44.00  & 47.81 
  & 59.81  & 75.00  & 64.67  \\

w/ Reflexion~\citep{shinn2023reflexion}
  & Trajectory
  & 77.59  & 73.68  & 76.04 
  & 51.97  & 43.47  & 48.65 
  & 61.74  & 72.00  & 64.61  \\

w/ Expel~\citep{zhao2024expel}
  & Trajectory
  & 81.03  & 76.32  & 79.17 
  & 53.68  & 42.93  & 49.48 
  & 59.16  & 76.00  & 65.95  \\

\rowcolor{ourslavender}
w/ MuSEAgent
  & State
  & \best{84.48} & \best{78.95} & \best{82.29}
  & \best{54.87} & \best{46.40} & \best{51.56}
  & \best{67.20} & \best{78.00} & \best{69.76} \\
\bottomrule
\end{tabular}
\end{adjustbox}
\label{tab:stateful_exp_learning}
\end{table*}

\subsection{Overall Performance}
\label{sec:performance}

Table~\ref{tab:stateful_exp_learning} presents the main results across various methods and base models. 
From these results, we derive two key findings:

\paragraph{Stateful experiences consistently outperform trajectory-level baselines.} MuSEAgent achieves the highest average accuracy across all evaluated benchmarks and base models. Compared with trajectory-level methods such as Reflexion and Expel, MuSEAgent shows substantial gains. For example, on Qwen3-VL-32B-Instruct, MuSEAgent reaches 65.30\% average accuracy, surpassing the strongest baseline by nearly 8\%. On the fine-grained V*~Bench Relative Position task with Qwen3-VL-235B-A22B-Instruct, our method improves accuracy by 18.43\% over Expel. These results suggest that decoupling historical episodes into granular state-action pairs mitigates the noise of monolithic trajectories. By retrieving precise state-aware experiences, the agent resolves multimodal reasoning bottlenecks more effectively than relying on coarse task-level analogies.

\paragraph{Stateful experiences benefit compact models most, while absolute performance scales with model size.} 
The absolute performance of MuSEAgent increases with model size, reaching 69.76\% on Qwen3.5-397B-A17B. However, relative improvements over trajectory-based baselines are larger for smaller models. MuSEAgent improves the best baseline by 7.97\% on Qwen3-VL-32B-Instruct, while gains on Qwen3-VL-235B-A22B-Instruct and Qwen3.5-397B-A17B are 3.25\% and 3.81\%, respectively. This pattern indicates that compact models, with limited intrinsic reasoning capacity, benefit more from fine-grained stateful experiences for complex multimodal tasks. Meanwhile, the consistent gains on Qwen3.5-397B-A17B show that targeted experience retrieval remains valuable even for large models.

\begin{figure*}[t]
    \centering
    \begin{minipage}[t]{0.49\textwidth}
        \centering
        \includegraphics[width=\linewidth]{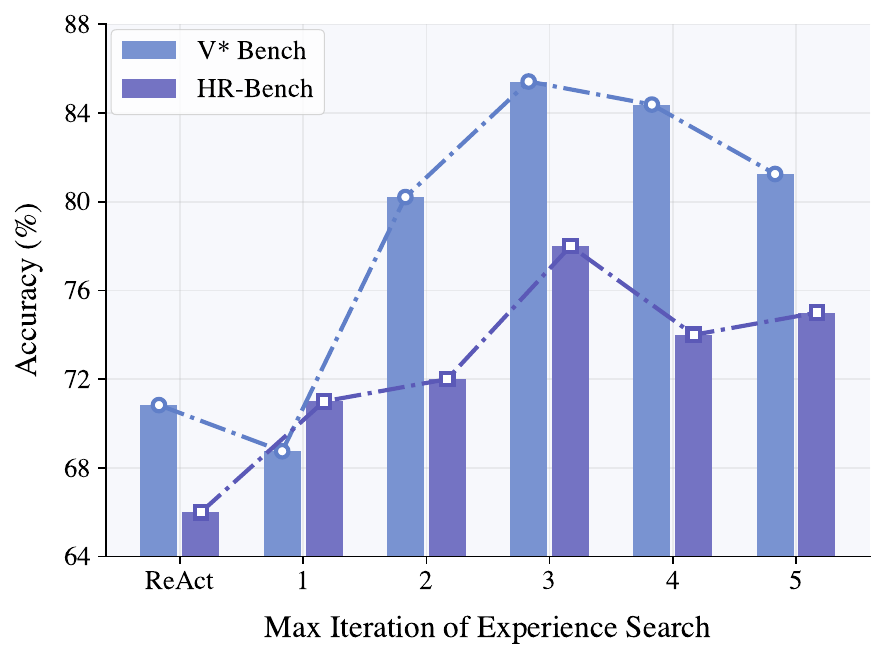}
        \caption*{(a) Deep Search}
    \end{minipage}
    \hfill
    \begin{minipage}[t]{0.49\textwidth}
        \centering
        \includegraphics[width=\linewidth]{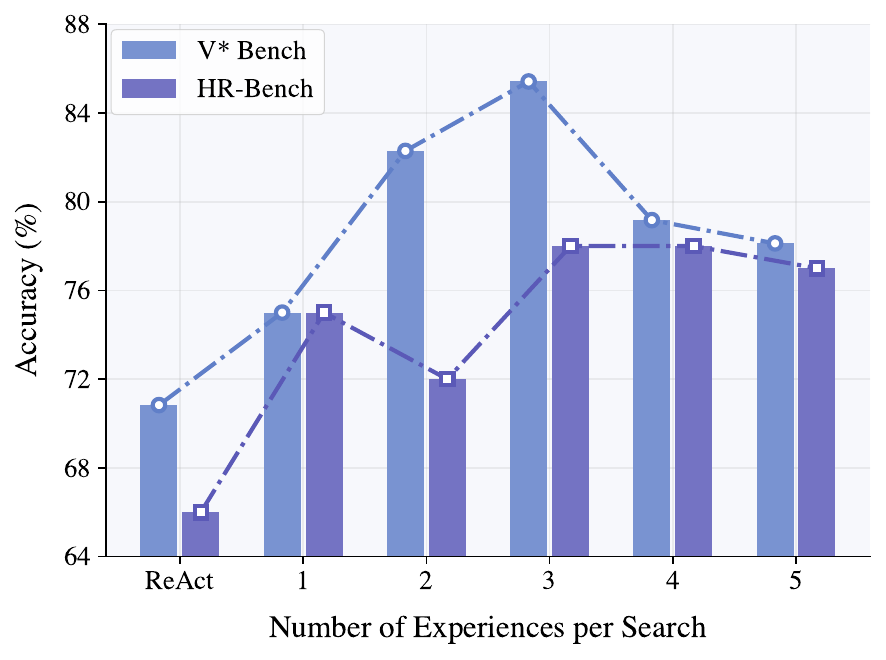}
        \caption*{(b) Wide Search}
    \end{minipage}
    \caption{Scaling behavior of the Deep-and-Wide search mechanism. We evaluate model performance by scaling (a) the maximum iterations of experience search (Deep Search) and (b) the number of experiences per search (Wide Search), with the alternate dimension fixed at 3 in each setting.}
    \label{fig:deep_wide_ablation}
\end{figure*}

\subsection{Investigation on Deep-and-Wide Experience Search}
\label{sec:search}

To evaluate the Deep-and-Wide search mechanism, we conduct controlled scaling experiments on the Qwen3-VL-32B-Instruct base model. During evaluation, we scale one retrieval dimension while fixing the other at 3. This setup isolates Deep Search by varying the maximum search iterations and Wide Search by scaling the number of retrieved experiences per search.

\paragraph{Scaling experience search depth and breadth consistently improves reasoning performance over the experience-free baseline.} Figure~\ref{fig:deep_wide_ablation} shows that incorporating the experience search mechanism improves agent performance across all configurations compared with the zero-experience ReAct baseline. For Deep Search, increasing the maximum search iterations from 0 to 3 raises accuracy from about 71\% to 85\% on V*~Bench and from 66\% to 78\% on HR-Bench. Wide Search shows a similar trend when scaling retrieved experiences per search from 1 to 3. Even when the retrieval parameters increase to 4 or 5, performance remains higher than the ReAct baseline, confirming that querying diverse semantic viewpoints provides effective state-aware guidance.

\paragraph{Performance achieves an optimal peak at moderate search scales before slightly declining.} Figure~\ref{fig:deep_wide_ablation} shows that reasoning performance peaks at 3 iterations for Deep Search and 3 retrieved experiences for Wide Search. Scaling either dimension to 4 or 5 leads to a slight drop, though still outperforming the baseline. This suggests that excessive historical experiences introduce redundant information into the context window, mildly diluting guidance for the current state. Consequently, maintaining a balanced search scope optimizes experience utilization without overwhelming the base model's instruction-following capacity.

\begin{table*}[t]
\caption{Generalization of OOD Stateful Experiences. We evaluate whether stateful experiences generalize well to unseen domains. Unlike baselines that use in-domain experiences, our method uses OOD experiences sourced from the other three datasets. We report the accuracy (\%) for all tasks. Best results are highlighted in \best{bold}.}
\centering
\fontsize{8pt}{10pt}\selectfont
\setlength{\tabcolsep}{3pt}

\begin{adjustbox}{max width=\textwidth}
\begin{tabular}{
  >{\raggedright\arraybackslash}l
  c
  c c c
  c c c
  c c c
}
\toprule
\multirow{2}{*}{\textbf{Method}} &
\multirow{2}{*}{\textbf{Exp. Level}} &
\multicolumn{3}{c}{\textbf{V* Bench}} &
\multicolumn{3}{c}{\textbf{MME-RealWorld-Lite}} &
\multirow{2}{*}{\textbf{Zoom-Bench}} &
\multirow{2}{*}{\textbf{HR-Bench}} &
\multirow{2}{*}{\textbf{Avg.}} \\
\cmidrule(lr){3-5}\cmidrule(lr){6-8}
& &
\textbf{Direct Attr.} & \textbf{Rel. Pos.} & \textbf{Micro Avg.}
& \textbf{Percep.} & \textbf{Reas.} & \textbf{Micro Avg.}
& & & \\
\midrule

\rowcolor{dividergray}
\multicolumn{11}{c}{\qwenicon \textbf{\textit{Qwen3-VL-32B-Instruct}}} \\

Vanilla CoT~\citep{kojima2022large}
& N/A
& 27.59  & 57.89  & 39.58 
& 28.55  & 28.53  & 28.54 
& 31.83  & 51.00  & 37.74  \\

w/ ReAct~\citep{yao2022react}
& N/A
& 62.07  & \best{84.21}  & 70.83 
& 44.44  & 36.80  & 41.46 
& 44.69  & 66.00  & 55.75  \\

w/ Reflexion~\citep{shinn2023reflexion}
  & Trajectory
  & {74.13} & {76.32} & {75.00}
  & {\best{46.15}} & {34.13} & {41.46}
  & {48.87} & {64.00} & {57.33} \\

w/ Expel~\citep{zhao2024expel}
& Trajectory
& 70.69  & 78.95  & 73.96 
& 42.39  & \best{40.27} & \best{41.56}
& 46.30  & 66.00  & 56.96  \\

\rowcolor{ourslavender}
w/ MuSEAgent (OOD)
& State
& \best{77.59} & 78.95  & \best{78.12}
& 42.56  & 36.00  & 40.00 
& \best{51.45} & \best{69.00} & \best{59.64} \\

\bottomrule
\end{tabular}
\end{adjustbox}
\label{tab:ood_stateful_exp}
\end{table*}

\subsection{Investigation on OOD Experience Generalization}
\label{sec:ood}

To evaluate the transferability of stateful experiences across unseen domains, we conduct OOD generalization experiments using the Qwen3-VL-32B-Instruct model. 
Following the optimal configuration in Sec.~\ref{sec:deepsearch}, we construct the experience bank for each target dataset using exploration records from the remaining three datasets, ensuring zero in-domain data exposure during reasoning.

\begin{figure}[t]
    \centering
    \includegraphics[width=0.8\linewidth]{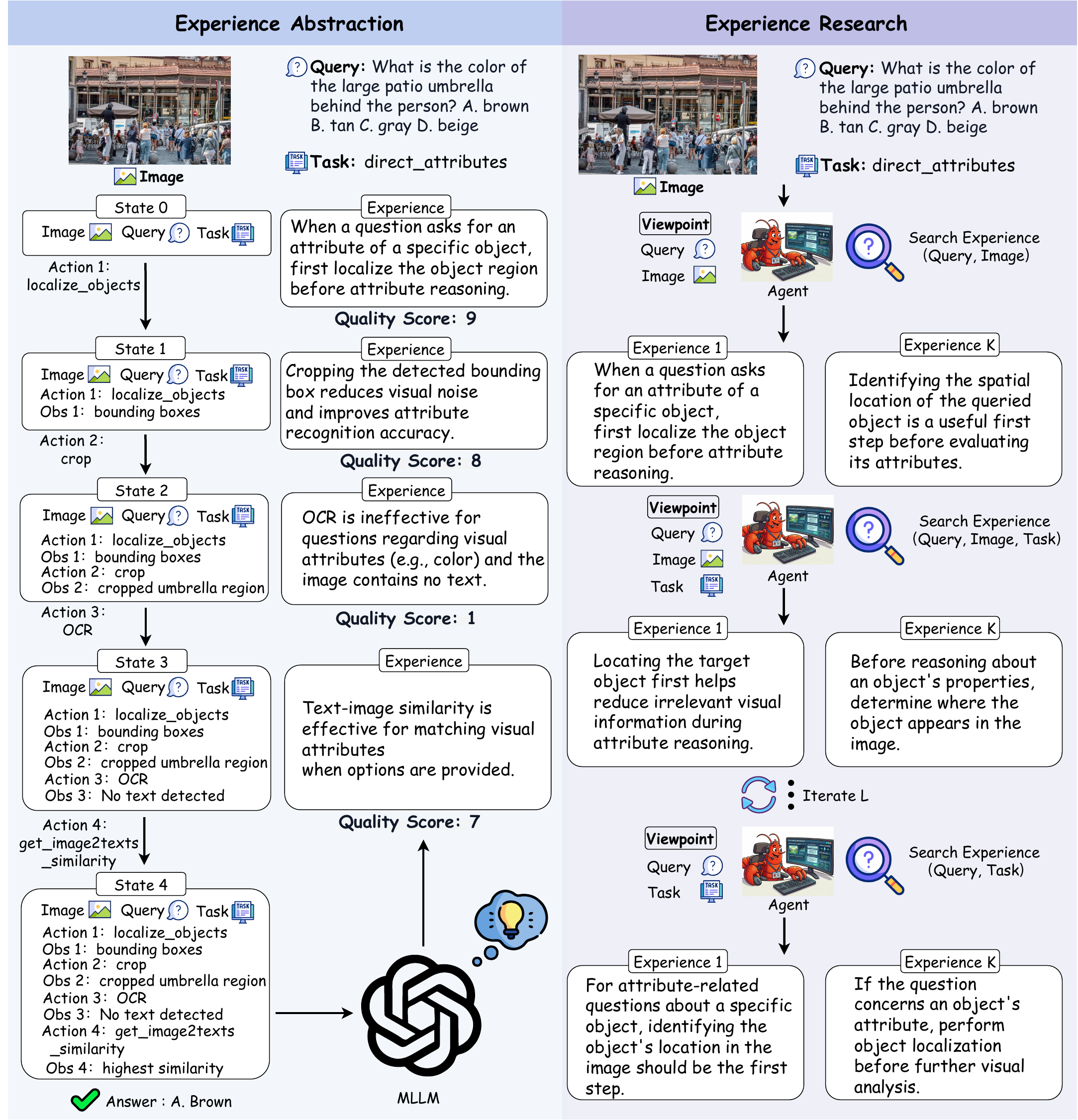}
    \caption{A Case of Stateful Experience Abstraction and Deep-and-Wide Search on Stateful Experiences.}
    \label{fig:case}
\end{figure}

\paragraph{OOD stateful experiences consistently surpass in-domain trajectory-level baselines.} 
Table~\ref{tab:ood_stateful_exp} shows that MuSEAgent using only OOD experiences achieves an average accuracy of 59.64\%, outperforming Reflexion (57.33\%) and Expel (56.96\%), both of which rely on target-domain trajectory data. This advantage is also evident on benchmarks requiring complex visual operations. For example, on Zoom-Bench and HR-Bench, MuSEAgent reaches 51.45\% and 69.00\%, exceeding the strongest in-domain baselines by 2.58\% and 3.00\% in absolute accuracy, respectively. These results suggest that decomposing trajectories into atomic state-action pairs captures transferable multimodal reasoning skills, enabling agents to apply learned strategies to unseen domains rather than relying on task-specific trajectory memorization.

\paragraph{Stateful experiences generalize across most tasks, though specialized domains still benefit from in-domain guidance.} 
While MuSEAgent transfers effectively on V* Bench, Zoom-Bench, and HR-Bench, performance on MME-RealWorld-Lite (40.00\%) is slightly lower than Expel (41.56\%) and the zero-experience ReAct baseline (41.46\%). This suggests that tasks relying on distinctive real-world visual characteristics may still benefit from target-domain experience. Nevertheless, the overall superiority of the OOD setting indicates that stateful experiences largely capture generalizable multimodal reasoning strategies, enabling robust cross-domain adaptation without prior exploration.

\begin{table*}[t]
\caption{Ablation of experience sources for MuSEAgent. We compare the impact of using correct, incorrect, or both types of trajectories as the experience source. We report the accuracy (\%) for all tasks. Best results are highlighted in \best{bold}.}
\centering
\fontsize{8pt}{10pt}\selectfont
\setlength{\tabcolsep}{3pt}

\begin{adjustbox}{max width=\textwidth}
\begin{tabular}{
  >{\raggedright\arraybackslash}l
  c
  c
  c c c c
}
\toprule
\multirow{2}{*}{\textbf{Method}} &
\multirow{2}{*}{\textbf{Exp. Level}} &
\multirow{2}{*}{\textbf{Exp. Source}} &
\multicolumn{3}{c}{\textbf{V* Bench}} &
\multirow{2}{*}{\textbf{HR-Bench}} \\
\cmidrule(lr){4-6}
& & &
\textbf{Direct Attr.} & \textbf{Rel. Pos.} & \textbf{Micro Avg.} &
\\
\midrule

\rowcolor{dividergray}
\multicolumn{7}{c}{\qwenicon \textbf{\textit{Qwen3-VL-32B-Instruct}}} \\

Vanilla CoT~\citep{kojima2022large}
& N/A
& N/A
& 27.59  & 57.89  & 39.58  & 51.00  \\

w/ ReAct~\citep{yao2022react}
& N/A
& N/A
& 62.07  & 84.21  & 70.83  & 66.00  \\

w/ Reflexion~\citep{shinn2023reflexion}
& Trajectory
& N/A
& 74.13  & 76.32  & 75.00  & 64.00  \\

w/ Expel~\citep{zhao2024expel}
& Trajectory
& N/A
& 70.69  & 78.95  & 73.96  & 66.00  \\

\rowcolor{ourslavender}
\textbf{}
& \textbf{}
& Correct Trajectories
& 62.07  & \best{86.84} & 71.88  & 72.00  \\

\rowcolor{ourslavender}
w/ MuSEAgent
& State
& Incorrect Trajectories
& 75.86  & 76.32  & 76.04  & 76.00  \\

\rowcolor{ourslavender}
\textbf{}
& \textbf{}
& Both
& \best{86.21} & 84.21  & \best{85.42} & \best{78.00} \\

\bottomrule
\end{tabular}
\end{adjustbox}
\label{tab:exp_source_ablation}
\end{table*}

\subsection{Ablation Studies}
\label{sec:ablation}

We conduct ablation studies on V*~Bench and HR-Bench using the Qwen3-VL-32B-Instruct model to analyze the effects of experience sources, hindsight reasoning models, and quality score thresholds.

\begin{table*}[t]
\caption{Ablation of hindsight models for MuSEAgent. We compare the impact of using different MLLMs (e.g., Qwen3-VL series and GPT-4o) for hindsight experience abstraction. 
}
\centering
\fontsize{8pt}{10pt}\selectfont
\setlength{\tabcolsep}{3pt}

\begin{adjustbox}{max width=\textwidth}
\begin{tabular}{
  >{\raggedright\arraybackslash}l
  c
  c
  c c c c
}
\toprule
\multirow{2}{*}{\textbf{Method}} &
\multirow{2}{*}{\textbf{Exp. Level}} &
\multirow{2}{*}{\textbf{HindSight Model}} &
\multicolumn{3}{c}{\textbf{V* Bench}} &
\multirow{2}{*}{\textbf{HR-Bench}} \\
\cmidrule(lr){4-6}
& & &
\textbf{Direct Attr.} & \textbf{Rel. Pos.} & \textbf{Micro Avg.} &
\\
\midrule

\rowcolor{dividergray}
\multicolumn{7}{c}{\qwenicon \textbf{\textit{Qwen3-VL-32B-Instruct}}} \\

Vanilla CoT~\citep{kojima2022large}
& N/A
& N/A
& 27.59  & 57.89  & 39.58  & 51.00  \\

w/ ReAct~\citep{yao2022react}
& N/A
& N/A
& 62.07  & 84.21  & 70.83  & 66.00  \\

w/ Reflexion~\citep{shinn2023reflexion}
& Trajectory
& N/A
& 74.13  & 76.32  & 75.00  & 64.00  \\

w/ Expel~\citep{zhao2024expel}
& Trajectory
& N/A
& 70.69  & 78.95  & 73.96  & 66.00  \\

\rowcolor{ourslavender}
& State
& \qwenicon Qwen3-VL-32B-Instruct
& 75.86  & 81.58  & 78.12  & 75.00  \\

\rowcolor{ourslavender}
w/Ours
& State
& \qwenicon Qwen3-VL-235B-A22B-Instruct
& 74.14  & 81.58  & 77.08  & 76.00  \\

\rowcolor{ourslavender}
& State
& \qwenicon Qwen3.5-397B-A17B
& 77.59  & \best{92.11} & 83.33  & 73.00  \\

\rowcolor{ourslavender}
& State
& \gpticon GPT-4o
& \best{86.21} & 84.21  & \best{85.42} & \best{78.00} \\

\bottomrule
\end{tabular}
\end{adjustbox}
\label{tab:hindsight_model_ablation}
\end{table*}

\paragraph{Experience Source.}
We first examine how different trajectory sources affect experience quality. Table~\ref{tab:exp_source_ablation} shows that using only successful trajectories achieves 71.88\% on V* Bench and 72.00\% on HR-Bench. Extracting experiences solely from failed interactions increases performance to 76.04\% and 76.00\%, suggesting that correcting erroneous reasoning provides stronger guidance than reinforcing successful paths. Combining both sources yields the best results, reaching 85.42\% on V* Bench and 78.00\% on HR-Bench. This indicates that a diverse experience bank enables the agent to both learn optimal strategies and avoid known pitfalls.

\paragraph{Hindsight Reasoning Model.}
Next, we analyze how the hindsight reasoning model affects experience quality. As shown in Table~\ref{tab:hindsight_model_ablation}, using Qwen3-VL-32B-Instruct for self-reflection achieves 78.12\% on V* Bench and 75.00\% on HR-Bench, already surpassing trajectory-level baselines. Larger models further improve performance: Qwen3.5-397B-A17B increases V* Bench accuracy to 83.33\%, while GPT-4o achieves the best results of 85.42\% and 78.00\% on the two benchmarks. This trend suggests that stronger multimodal models generate more precise state-aware experiences, leading to a higher-quality experience bank.

\begin{table*}[t]
\caption{Ablation of quality score thresholds for MuSEAgent. We report the accuracy (\%) for all tasks. Best results are highlighted in \best{bold}.}
\centering
\fontsize{8pt}{10pt}\selectfont
\setlength{\tabcolsep}{3pt}

\begin{adjustbox}{max width=\textwidth}
\begin{tabular}{
  >{\raggedright\arraybackslash}l
  c
  c
  c c c
  c
}
\toprule
\multirow{2}{*}{\textbf{Method}} &
\multirow{2}{*}{\textbf{Exp. Level}} &
\multirow{2}{*}{\textbf{Quality Score Threshold}} &
\multicolumn{3}{c}{\textbf{V* Bench}} &
\multirow{2}{*}{\textbf{HR-Bench}} \\
\cmidrule(lr){4-6}
& & &
\textbf{Direct Attr.} & \textbf{Rel. Pos.} & \textbf{Micro Avg.} &
\\
\midrule

\rowcolor{dividergray}
\multicolumn{7}{c}{\qwenicon \textbf{\textit{Qwen3-VL-32B-Instruct}}} \\

Direct CoT~\citep{kojima2022large}
& N/A
& N/A
& 27.59  & 57.89  & 39.58  & 51.00  \\

w/ ReAct~\citep{yao2022react}
& N/A
& N/A
& 62.07  & 84.21  & 70.83  & 66.00  \\

w/ Reflexion~\citep{shinn2023reflexion}
& Trajectory
& N/A
& 74.13  & 76.32  & 75.00  & 64.00  \\

w/ Expel~\citep{zhao2024expel}
& Trajectory
& N/A
& 70.69  & 78.95  & 73.96  & 66.00  \\

\rowcolor{ourslavender}
& State
& 9.0
& 72.41  & 78.95  & 75.00  & 73.00  \\

\rowcolor{ourslavender}
w/ MuSEAgent
& State
& 7.0
& 77.59  & 78.95  & 78.12  & 74.00  \\

\rowcolor{ourslavender}
& State
& 5.0
& \best{86.21} & \best{84.21} & \best{85.42} & \best{78.00} \\

\bottomrule
\end{tabular}
\end{adjustbox}
\label{tab:qvalue_threshold_ablation}
\end{table*}

\paragraph{Quality Score Threshold.}
Finally, we evaluate the effect of the quality score threshold for filtering experiences. Table~\ref{tab:qvalue_threshold_ablation} shows that a strict threshold of 9.0 limits the bank to near-perfect experiences, yielding 75.00\% accuracy on V* Bench and 73.00\% on HR-Bench. Relaxing the threshold to 7.0 improves accuracy to 78.12\% and 74\%, while further reducing it to 5.0 achieves the best results of 85.42\% on V* Bench and 78.00\% on HR-Bench. This indicates that overly strict filtering removes valuable partial successes and nuanced strategic experience, whereas retaining moderately scored experiences provides richer guidance for diverse multimodal states.

\subsection{Case Study}
\label{sec:case_study}

We further provide a case of our stateful experience abstraction process and Deep-and-Wide search process in Figure~\ref{fig:case}. The left panel illustrates a historical interaction trajectory for a fine-grained attribute recognition task, where the agent sequentially executes object localization, image cropping, optical character recognition, and text-image similarity matching to determine the color of a specified umbrella. Instead of caching this entire monolithic sequence, the hindsight reasoning module evaluates each discrete state-action pair to extract granular strategic rules alongside corresponding quality scores. For instance, the initial object localization action receives a high quality score of 9, as identifying the target spatial region is a critical prerequisite for accurate attribute evaluation. Conversely, the optical character recognition operation applied to the cropped umbrella region yields a quality score of 1, explicitly documenting that text extraction is an ineffective strategy for queries concerning purely visual properties. This granular decoupling ensures that the experience bank retains precise, state-aware guidance while filtering out suboptimal or noisy intermediate steps.

Building upon the constructed stateful experience bank, the right panel demonstrates how the agent leverages compositional state representations to conduct active experience search during online inference. When confronted with a given multimodal state, the agent decomposes its current context into distinct semantic viewpoints, such as pairing the user query exclusively with the visual input or isolating the query alongside the formal task description. Utilizing these varying semantic viewpoints, the agent iteratively searches the experience bank to retrieve relevant historical insights. The retrieved results, despite originating from different semantic retrieval indices, consistently converge on aligned strategic advice emphasizing the necessity of spatial localization prior to detailed visual analysis. By dynamically aggregating these targeted insights through the iterative experience search mechanism, the agent synthesizes comprehensive guidance to formulate an optimal execution plan, thereby overcoming the structural rigidity inherent in traditional trajectory-level methods.

\section{Conclusion}
We propose MuSEAgent, a multimodal reasoning agent that mitigates the noise and inflexible context modeling of traditional trajectory-level experience retrieval by decoupling historical episodes into atomic state-action pairs. Utilizing hindsight reasoning, we construct a granular stateful experience bank that agents dynamically query through a Deep-and-Wide search paradigm. Empirical evaluations on various benchmarks demonstrate that this state-aware guidance consistently outperforms existing trajectory-based methods across diverse base model architectures. Furthermore, out-of-domain generalization experiments indicate that stateful experiences effectively capture generalizable reasoning and tool-use skills, enabling robust out-of-distribution transfer without prior in-domain exploration. Future research will explore online experience construction mechanisms, allowing agents to continuously update their stateful memory and autonomously self-improve during active interaction with open-ended environments.

\bibliographystyle{plainnat}
\bibliography{citation}

\begin{thebibliography}{37}
\providecommand{\natexlab}[1]{#1}
\providecommand{\url}[1]{\texttt{#1}}
\expandafter\ifx\csname urlstyle\endcsname\relax
  \providecommand{\doi}[1]{doi: #1}\else
  \providecommand{\doi}{doi: \begingroup \urlstyle{rm}\Url}\fi

\bibitem[Bai et~al.(2023)Bai, Bai, Chu, Cui, Dang, Deng, Fan, Ge, Han, Huang, et~al.]{bai2023qwen}
Jinze Bai, Shuai Bai, Yunfei Chu, Zeyu Cui, Kai Dang, Xiaodong Deng, Yang Fan, Wenbin Ge, Yu~Han, Fei Huang, et~al.
\newblock Qwen technical report.
\newblock \emph{arXiv preprint arXiv:2309.16609}, 2023.

\bibitem[Bai et~al.(2025)Bai, Cai, Chen, Chen, Chen, Cheng, Deng, Ding, Gao, Ge, Ge, Guo, Huang, Huang, Huang, Hui, Jiang, Li, Li, Li, Li, Lin, Lin, Liu, Liu, Liu, Liu, Liu, Liu, Lu, Luo, Lv, Men, Meng, Ren, Ren, Song, Sun, Tang, Tu, Wan, Wang, Wang, Wang, Wang, Xie, Xu, Xu, Xu, Yang, Yang, Yang, Yang, Yu, Zhang, Zhang, Zhang, Zheng, Zhong, Zhou, Zhou, Zhou, Zhu, and Zhu]{Qwen3-VL}
Shuai Bai, Yuxuan Cai, Ruizhe Chen, Keqin Chen, Xionghui Chen, Zesen Cheng, Lianghao Deng, Wei Ding, Chang Gao, Chunjiang Ge, Wenbin Ge, Zhifang Guo, Qidong Huang, Jie Huang, Fei Huang, Binyuan Hui, Shutong Jiang, Zhaohai Li, Mingsheng Li, Mei Li, Kaixin Li, Zicheng Lin, Junyang Lin, Xuejing Liu, Jiawei Liu, Chenglong Liu, Yang Liu, Dayiheng Liu, Shixuan Liu, Dunjie Lu, Ruilin Luo, Chenxu Lv, Rui Men, Lingchen Meng, Xuancheng Ren, Xingzhang Ren, Sibo Song, Yuchong Sun, Jun Tang, Jianhong Tu, Jianqiang Wan, Peng Wang, Pengfei Wang, Qiuyue Wang, Yuxuan Wang, Tianbao Xie, Yiheng Xu, Haiyang Xu, Jin Xu, Zhibo Yang, Mingkun Yang, Jianxin Yang, An~Yang, Bowen Yu, Fei Zhang, Hang Zhang, Xi~Zhang, Bo~Zheng, Humen Zhong, Jingren Zhou, Fan Zhou, Jing Zhou, Yuanzhi Zhu, and Ke~Zhu.
\newblock Qwen3-vl technical report.
\newblock \emph{arXiv preprint arXiv:2511.21631}, 2025.

\bibitem[Chang et~al.(2024)Chang, Zhang, Zhu, Yang, Yang, Jin, Lan, Kong, and He]{chang2024agentboard}
Ma~Chang, Junlei Zhang, Zhihao Zhu, Cheng Yang, Yujiu Yang, Yaohui Jin, Zhenzhong Lan, Lingpeng Kong, and Junxian He.
\newblock Agentboard: An analytical evaluation board of multi-turn llm agents.
\newblock \emph{Advances in neural information processing systems}, 37:\penalty0 74325--74362, 2024.

\bibitem[Dong et~al.(2025)Dong, Liu, Sun, Yang, Hu, Rao, and Liu]{dong2025insight}
Yuhao Dong, Zuyan Liu, Hai-Long Sun, Jingkang Yang, Winston Hu, Yongming Rao, and Ziwei Liu.
\newblock Insight-v: Exploring long-chain visual reasoning with multimodal large language models.
\newblock In \emph{Proceedings of the Computer Vision and Pattern Recognition Conference}, pages 9062--9072, 2025.

\bibitem[Gupta and Kembhavi(2023)]{gupta2023visual}
Tanmay Gupta and Aniruddha Kembhavi.
\newblock Visual programming: Compositional visual reasoning without training.
\newblock In \emph{Proceedings of the IEEE/CVF conference on computer vision and pattern recognition}, pages 14953--14962, 2023.

\bibitem[He et~al.(2024)He, Yao, Ma, Yu, Dai, Zhang, Lan, and Yu]{he2024webvoyager}
Hongliang He, Wenlin Yao, Kaixin Ma, Wenhao Yu, Yong Dai, Hongming Zhang, Zhenzhong Lan, and Dong Yu.
\newblock Webvoyager: Building an end-to-end web agent with large multimodal models.
\newblock In \emph{Proceedings of the 62nd Annual Meeting of the Association for Computational Linguistics (Volume 1: Long Papers)}, pages 6864--6890, 2024.

\bibitem[Hurst et~al.(2024)Hurst, Lerer, Goucher, Perelman, Ramesh, Clark, Ostrow, Welihinda, Hayes, Radford, et~al.]{hurst2024gpt}
Aaron Hurst, Adam Lerer, Adam~P Goucher, Adam Perelman, Aditya Ramesh, Aidan Clark, AJ~Ostrow, Akila Welihinda, Alan Hayes, Alec Radford, et~al.
\newblock Gpt-4o system card.
\newblock \emph{arXiv preprint arXiv:2410.21276}, 2024.

\bibitem[Kojima et~al.(2022)Kojima, Gu, Reid, Matsuo, and Iwasawa]{kojima2022large}
Takeshi Kojima, Shixiang~Shane Gu, Machel Reid, Yutaka Matsuo, and Yusuke Iwasawa.
\newblock Large language models are zero-shot reasoners.
\newblock \emph{Advances in neural information processing systems}, 35:\penalty0 22199--22213, 2022.

\bibitem[Li et~al.(2026)Li, Zhang, Long, Chen, Song, Bai, Yang, Xie, Yang, Liu, et~al.]{li2026qwen3}
Mingxin Li, Yanzhao Zhang, Dingkun Long, Keqin Chen, Sibo Song, Shuai Bai, Zhibo Yang, Pengjun Xie, An~Yang, Dayiheng Liu, et~al.
\newblock Qwen3-vl-embedding and qwen3-vl-reranker: A unified framework for state-of-the-art multimodal retrieval and ranking.
\newblock \emph{arXiv preprint arXiv:2601.04720}, 2026.

\bibitem[Lin and Xu(2025)]{lin2025understanding}
Heng Lin and Zhongwen Xu.
\newblock Understanding tool-integrated reasoning.
\newblock \emph{arXiv preprint arXiv:2508.19201}, 2025.

\bibitem[Liu et~al.(2023)Liu, Li, Wu, and Lee]{liu2023visual}
Haotian Liu, Chunyuan Li, Qingyang Wu, and Yong~Jae Lee.
\newblock Visual instruction tuning.
\newblock \emph{Advances in neural information processing systems}, 36:\penalty0 34892--34916, 2023.

\bibitem[Liu et~al.(2024{\natexlab{a}})Liu, Li, Li, and Lee]{liu2024improved}
Haotian Liu, Chunyuan Li, Yuheng Li, and Yong~Jae Lee.
\newblock Improved baselines with visual instruction tuning.
\newblock In \emph{Proceedings of the IEEE/CVF conference on computer vision and pattern recognition}, pages 26296--26306, 2024{\natexlab{a}}.

\bibitem[Liu et~al.(2024{\natexlab{b}})Liu, Li, Li, Li, Zhang, Shen, and Lee]{liu2024llavanext}
Haotian Liu, Chunyuan Li, Yuheng Li, Bo~Li, Yuanhan Zhang, Sheng Shen, and Yong~Jae Lee.
\newblock Llava-next: Improved reasoning, ocr, and world knowledge, January 2024{\natexlab{b}}.
\newblock URL \url{https://llava-vl.github.io/blog/2024-01-30-llava-next/}.

\bibitem[Liu et~al.(2024{\natexlab{c}})Liu, Lin, Hewitt, Paranjape, Bevilacqua, Petroni, and Liang]{liu2024lost}
Nelson~F Liu, Kevin Lin, John Hewitt, Ashwin Paranjape, Michele Bevilacqua, Fabio Petroni, and Percy Liang.
\newblock Lost in the middle: How language models use long contexts.
\newblock \emph{Transactions of the association for computational linguistics}, 12:\penalty0 157--173, 2024{\natexlab{c}}.

\bibitem[Packer et~al.(2023)Packer, Fang, Patil, Lin, Wooders, and Gonzalez]{packer2023memgpt}
Charles Packer, Vivian Fang, Shishir\_G Patil, Kevin Lin, Sarah Wooders, and Joseph\_E Gonzalez.
\newblock Memgpt: towards llms as operating systems.
\newblock 2023.

\bibitem[{Qwen Team}(2026)]{qwen3.5}
{Qwen Team}.
\newblock {Qwen3.5}: Towards native multimodal agents, February 2026.
\newblock URL \url{https://qwen.ai/blog?id=qwen3.5}.

\bibitem[Shen et~al.(2023)Shen, Song, Tan, Li, Lu, and Zhuang]{shen2023hugginggpt}
Yongliang Shen, Kaitao Song, Xu~Tan, Dongsheng Li, Weiming Lu, and Yueting Zhuang.
\newblock Hugginggpt: Solving ai tasks with chatgpt and its friends in hugging face.
\newblock \emph{Advances in Neural Information Processing Systems}, 36:\penalty0 38154--38180, 2023.

\bibitem[Shinn et~al.(2023)Shinn, Cassano, Gopinath, Narasimhan, and Yao]{shinn2023reflexion}
Noah Shinn, Federico Cassano, Ashwin Gopinath, Karthik Narasimhan, and Shunyu Yao.
\newblock Reflexion: Language agents with verbal reinforcement learning.
\newblock \emph{Advances in neural information processing systems}, 36:\penalty0 8634--8652, 2023.

\bibitem[Sur{\'\i}s et~al.(2023)Sur{\'\i}s, Menon, and Vondrick]{suris2023vipergpt}
D{\'\i}dac Sur{\'\i}s, Sachit Menon, and Carl Vondrick.
\newblock Vipergpt: Visual inference via python execution for reasoning.
\newblock In \emph{Proceedings of the IEEE/CVF international conference on computer vision}, pages 11888--11898, 2023.

\bibitem[Thawakar et~al.(2025)Thawakar, Dissanayake, More, Thawkar, Heakl, Ahsan, Li, Zumri, Lahoud, Anwer, et~al.]{thawakar2025llamav}
Omkar Thawakar, Dinura Dissanayake, Ketan~Pravin More, Ritesh Thawkar, Ahmed Heakl, Noor Ahsan, Yuhao Li, Ilmuz Zaman~Mohammed Zumri, Jean Lahoud, Rao~Muhammad Anwer, et~al.
\newblock Llamav-o1: Rethinking step-by-step visual reasoning in llms.
\newblock In \emph{Findings of the Association for Computational Linguistics: ACL 2025}, pages 24290--24315, 2025.

\bibitem[Wang et~al.(2023)Wang, Xie, Jiang, Mandlekar, Xiao, Zhu, Fan, and Anandkumar]{wang2023voyager}
Guanzhi Wang, Yuqi Xie, Yunfan Jiang, Ajay Mandlekar, Chaowei Xiao, Yuke Zhu, Linxi Fan, and Anima Anandkumar.
\newblock Voyager: An open-ended embodied agent with large language models.
\newblock \emph{arXiv preprint arXiv:2305.16291}, 2023.

\bibitem[Wang(2025)]{wang2025memento}
Jun Wang.
\newblock Memento-ii: Learning by stateful reflective memory.
\newblock \emph{arXiv preprint arXiv:2512.22716}, 2025.

\bibitem[Wang et~al.(2024)Wang, Ma, Feng, Zhang, Yang, Zhang, Chen, Tang, Chen, Lin, et~al.]{wang2024survey}
Lei Wang, Chen Ma, Xueyang Feng, Zeyu Zhang, Hao Yang, Jingsen Zhang, Zhiyuan Chen, Jiakai Tang, Xu~Chen, Yankai Lin, et~al.
\newblock A survey on large language model based autonomous agents.
\newblock \emph{Frontiers of Computer Science}, 18\penalty0 (6):\penalty0 186345, 2024.

\bibitem[Wang et~al.(2025)Wang, Ding, Zeng, Zhou, Shen, Luo, Yu, and Tao]{wang2025divide}
Wenbin Wang, Liang Ding, Minyan Zeng, Xiabin Zhou, Li~Shen, Yong Luo, Wei Yu, and Dacheng Tao.
\newblock Divide, conquer and combine: A training-free framework for high-resolution image perception in multimodal large language models.
\newblock In \emph{Proceedings of the AAAI Conference on Artificial Intelligence}, volume~39, pages 7907--7915, 2025.

\bibitem[Wei et~al.(2026)Wei, He, Lan, Dong, Cai, Li, Zhu, Wang, Kong, Wang, Zhang, and Huang]{wei2026zooming}
Lai Wei, Liangbo He, Jun Lan, Lingzhong Dong, Yutong Cai, Siyuan Li, Huijia Zhu, Weiqiang Wang, Linghe Kong, Yue Wang, Zhuosheng Zhang, and Weiran Huang.
\newblock Zooming without zooming: Region-to-image distillation for fine-grained multimodal perception.
\newblock \emph{arXiv preprint arXiv:2602.11858}, 2026.

\bibitem[Wu and Xie(2024)]{wu2024v}
Penghao Wu and Saining Xie.
\newblock V?: Guided visual search as a core mechanism in multimodal llms.
\newblock In \emph{Proceedings of the IEEE/CVF Conference on Computer Vision and Pattern Recognition}, pages 13084--13094, 2024.

\bibitem[Xi et~al.(2025)Xi, Chen, Guo, He, Ding, Hong, Zhang, Wang, Jin, Zhou, et~al.]{xi2025rise}
Zhiheng Xi, Wenxiang Chen, Xin Guo, Wei He, Yiwen Ding, Boyang Hong, Ming Zhang, Junzhe Wang, Senjie Jin, Enyu Zhou, et~al.
\newblock The rise and potential of large language model based agents: A survey.
\newblock \emph{Science China Information Sciences}, 68\penalty0 (2):\penalty0 121101, 2025.

\bibitem[Xie et~al.(2024)Xie, Zhang, Chen, Li, Zhao, Cao, Hua, Cheng, Shin, Lei, et~al.]{xie2024osworld}
Tianbao Xie, Danyang Zhang, Jixuan Chen, Xiaochuan Li, Siheng Zhao, Ruisheng Cao, Toh~J Hua, Zhoujun Cheng, Dongchan Shin, Fangyu Lei, et~al.
\newblock Osworld: Benchmarking multimodal agents for open-ended tasks in real computer environments.
\newblock \emph{Advances in Neural Information Processing Systems}, 37:\penalty0 52040--52094, 2024.

\bibitem[Xu et~al.(2025)Xu, Jin, Wu, Li, Song, Sun, and Yuan]{xu2025llava}
Guowei Xu, Peng Jin, Ziang Wu, Hao Li, Yibing Song, Lichao Sun, and Li~Yuan.
\newblock Llava-cot: Let vision language models reason step-by-step.
\newblock In \emph{Proceedings of the IEEE/CVF International Conference on Computer Vision}, pages 2087--2098, 2025.

\bibitem[Yang et~al.(2023)Yang, Li, Wang, Lin, Azarnasab, Ahmed, Liu, Liu, Zeng, and Wang]{yang2023mm}
Zhengyuan Yang, Linjie Li, Jianfeng Wang, Kevin Lin, Ehsan Azarnasab, Faisal Ahmed, Zicheng Liu, Ce~Liu, Michael Zeng, and Lijuan Wang.
\newblock Mm-react: Prompting chatgpt for multimodal reasoning and action.
\newblock \emph{arXiv preprint arXiv:2303.11381}, 2023.

\bibitem[Yao et~al.(2022)Yao, Zhao, Yu, Du, Shafran, Narasimhan, and Cao]{yao2022react}
Shunyu Yao, Jeffrey Zhao, Dian Yu, Nan Du, Izhak Shafran, Karthik~R Narasimhan, and Yuan Cao.
\newblock React: Synergizing reasoning and acting in language models.
\newblock In \emph{The eleventh international conference on learning representations}, 2022.

\bibitem[Yao et~al.(2023)Yao, Heinecke, Niebles, Liu, Feng, Xue, Murthy, Chen, Zhang, Arpit, et~al.]{yao2023retroformer}
Weiran Yao, Shelby Heinecke, Juan~Carlos Niebles, Zhiwei Liu, Yihao Feng, Le~Xue, Rithesh Murthy, Zeyuan Chen, Jianguo Zhang, Devansh Arpit, et~al.
\newblock Retroformer: Retrospective large language agents with policy gradient optimization.
\newblock \emph{arXiv preprint arXiv:2308.02151}, 2023.

\bibitem[Zhang et~al.(2025)Zhang, Yang, Liu, Li, Han, Chen, Huang, Fu, and Yu]{zhang2025appagent}
Chi Zhang, Zhao Yang, Jiaxuan Liu, Yanda Li, Yucheng Han, Xin Chen, Zebiao Huang, Bin Fu, and Gang Yu.
\newblock Appagent: Multimodal agents as smartphone users.
\newblock In \emph{Proceedings of the 2025 CHI Conference on Human Factors in Computing Systems}, pages 1--20, 2025.

\bibitem[Zhang et~al.(2024)Zhang, Zhang, Tian, Fu, Zhang, Wu, Li, Wang, Wen, Zhang, et~al.]{zhang2024mme}
Yi-Fan Zhang, Huanyu Zhang, Haochen Tian, Chaoyou Fu, Shuangqing Zhang, Junfei Wu, Feng Li, Kun Wang, Qingsong Wen, Zhang Zhang, et~al.
\newblock Mme-realworld: Could your multimodal llm challenge high-resolution real-world scenarios that are difficult for humans?
\newblock \emph{arXiv preprint arXiv:2408.13257}, 2024.

\bibitem[Zhao et~al.(2024)Zhao, Huang, Xu, Lin, Liu, and Huang]{zhao2024expel}
Andrew Zhao, Daniel Huang, Quentin Xu, Matthieu Lin, Yong-Jin Liu, and Gao Huang.
\newblock Expel: Llm agents are experiential learners.
\newblock In \emph{Proceedings of the AAAI Conference on Artificial Intelligence}, volume~38, pages 19632--19642, 2024.

\bibitem[Zhou et~al.(2025)Zhou, Chen, Guo, Yan, Lee, Wang, Lee, Zhang, Shao, Yang, et~al.]{zhou2025memento}
Huichi Zhou, Yihang Chen, Siyuan Guo, Xue Yan, Kin~Hei Lee, Zihan Wang, Ka~Yiu Lee, Guchun Zhang, Kun Shao, Linyi Yang, et~al.
\newblock Memento: Fine-tuning llm agents without fine-tuning llms.
\newblock \emph{arXiv preprint arXiv:2508.16153}, 2025.

\bibitem[Zhu et~al.(2023)Zhu, Chen, Tian, Tao, Su, Yang, Huang, Li, Lu, Wang, et~al.]{zhu2023ghost}
Xizhou Zhu, Yuntao Chen, Hao Tian, Chenxin Tao, Weijie Su, Chenyu Yang, Gao Huang, Bin Li, Lewei Lu, Xiaogang Wang, et~al.
\newblock Ghost in the minecraft: Generally capable agents for open-world environments via large language models with text-based knowledge and memory.
\newblock \emph{arXiv preprint arXiv:2305.17144}, 2023.

\end{thebibliography}

\clearpage
\appendix

\lstset{numbers=none}

\section*{Appendix}

\section{Datasets and Metrics}
\label{app:dataset}
We summarize the benchmarks and corresponding metrics used to evaluate the fine-grained perception and high-resolution reasoning capabilities of multimodal models below:

\paragraph{V* Bench~\citep{wu2024v}.} 
This Visual Question Answering (VQA) benchmark is designed to evaluate the detailed visual grounding capabilities of multimodal models on high-resolution images. It consists of 191 carefully curated samples divided into two fine-grained sub-tasks: attribute recognition and spatial relationship reasoning. All questions are presented in a multiple-choice format. The benchmark tests whether models can accurately perceive small, critical visual details that standard vision encoders typically overlook. We use the overall accuracy across these sub-tasks as the final score.

\paragraph{MME-RealWorld-Lite~\citep{zhang2024mme}.} 
This is an accelerated evaluation subset of the large-scale MME-RealWorld benchmark, which tests models on high-resolution images across challenging, practical real-world scenarios. By efficiently subsampling up to 50 instances across 43 tasks, the Lite version maintains diverse scenario coverage with reduced computational cost. For our evaluation, we select the multiple-choice questions from this dataset, totaling 1,919 items. We use the average task accuracy as the final score.

\paragraph{Zoom-Bench~\citep{wei2026zooming}.}
Consisting of 845 instances, this VQA benchmark is designed to assess fine-grained multimodal perception. It specifically focuses on challenging scenarios where the decisive visual evidence is extremely small and easily overshadowed by the global context, evaluating a model's ability to accurately capture minute details. In our evaluation, we select the multiple-choice subset, which contains 621 questions. We use the overall VQA accuracy as the final score.

\paragraph{HR-Bench~\citep{wang2025divide}.}
This benchmark is deliberately constructed to measure multimodal models' perception and reasoning capabilities on true high-resolution images. It focuses on fine-grained single-instance and cross-instance perception tasks, systematically evaluating how well models extract and reason over intricate visual details typically lost during standard image down-sampling. In our evaluation, we select 200 multiple-choice questions based on 8K-resolution images from HR-Bench, and use the overall accuracy on these high-resolution questions as the final score.

\section{Baseline Methods}
\label{app:baselines}
The baseline methods we compare against in our experiments are summarized as follows:

\paragraph{Vanilla CoT~\citep{kojima2022large}.} 
Chain-of-Thought prompting directs LLMs/VLMs to generate a sequence of intermediate reasoning steps before producing a final answer. This method relies solely on the model's pre-trained parametric knowledge to decompose complex problems into sequential logical deductions. Operating without external tool invocations or environmental feedback, Vanilla CoT serves as a baseline to evaluate the intrinsic reasoning capacity and standalone problem-solving performance of language models.

\paragraph{ReAct~\citep{yao2022react}.} This framework integrates reasoning and acting within LLMs/VLMs. Instead of relying solely on internal knowledge, ReAct enables models to generate reasoning traces and task-specific actions in an interleaved manner. This process allows the agent to formulate and adjust plans while interacting with external environments, such as knowledge bases or APIs, to retrieve real-time observations. By grounding its reasoning in external information, ReAct provides a mechanism to reduce factual errors and maintain interpretability during multi-step decision-making tasks.

\paragraph{Reflexion~\citep{shinn2023reflexion}.}
Reflexion equips agents with a mechanism for improving their behavior through verbal self-reflection. After completing a task attempt, an evaluator analyzes the resulting trajectory and produces natural language reflections based on environmental feedback. These reflections are stored in an episodic memory buffer and provided as additional context in subsequent attempts, enabling the agent to revise its reasoning strategy and avoid repeating previous mistakes. In our implementation, reflective experiences are extracted only from failed trajectories.

\paragraph{Expel~\citep{zhao2024expel}.}
Expel is a method that enables agents to distill reusable knowledge from their interactions with the environment. By analyzing task trajectories, the agent extracts generalized rules or strategies that can guide future decision making. Unlike approaches that focus primarily on failure analysis, Expel leverages information from both successful and unsuccessful experiences to identify patterns that improve planning and reasoning. The extracted knowledge is stored in an experience pool and incorporated into future task executions. In our implementation, experiences are extracted from both successful and failed trajectories.

\section{Tool Descriptions}
\label{app:tool}
We equip our model with a diverse set of tools to interact with visual inputs, perform calculations, and access external information. These tools are categorized into four functional modules below.

\subsection{Image Manipulation Tools}

\paragraph{Zoom In.} This tool scales a specific region of an image based on a provided 2D bounding box and a numerical zoom factor. It outputs the magnified image segment to facilitate the inspection of minute details.

\begin{lstlisting}[language=Python]
class ZoomInTool(BasicTool):
    name = "zoom_in"
    description_en = "Zoom in on a region by cropping and upscaling. Use localize_objects first if bbox is unknown."
    
    parameters = {
        "type": "object",
        "properties": {
            "image": {
                "type": "string",
                "description": "Image ID (e.g., 'img_0')"
            },
            "bbox_2d": {
                "type": "array",
                "items": {"type": "number"},
                "minItems": 4,
                "maxItems": 4,
                "description": "Bounding box [left, top, right, bottom], values between 0 and 1"
            },
            "zoom_factor": {
                "type": "number",
                "description": "Zoom factor (must be > 1)",
                "exclusiveMinimum": 1.0
            }
        },
        "required": ["image", "bbox_2d", "zoom_factor"]
    }
    example = '{"image": "img_0", "bbox_2d": [0.25, 0.25, 0.75, 0.75], "zoom_factor": 2.0}'
    
    def call(self, params: Union[str, Dict]) -> Dict:
        params_dict = self.parse_params(params)
        
        image_path = params_dict["image"]
        bbox_2d = params_dict["bbox_2d"]
        zoom_factor = params_dict["zoom_factor"]
        
        if not all(0 <= x <= 1.0 for x in bbox_2d):
            return {"error": "bbox_2d values must be between 0 and 1"}
        if bbox_2d[0] >= bbox_2d[2] or bbox_2d[1] >= bbox_2d[3]:
            return {"error": "invalid bbox_2d: left >= right or top >= bottom"}
        if zoom_factor <= 1:
            return {"error": "zoom_factor must be > 1"}
        
        try:
            image = image_processing(image_path)
        except Exception as e:
            return {"error": f"failed to load image: {e}"}
        
        W, H = image.size
        crop_box = (int(bbox_2d[0] * W), int(bbox_2d[1] * H), int(bbox_2d[2] * W), int(bbox_2d[3] * H))
        cropped = image.crop(crop_box)
        
        new_size = (int(cropped.width * zoom_factor), int(cropped.height * zoom_factor))
        zoomed = cropped.resize(new_size, Image.LANCZOS)
        
        return {"output_image": zoomed}
    
    def generate_description(self, properties, observation):
        img = properties.get("image", "image")
        bbox_2d = properties.get("bbox_2d", [])
        factor = properties.get("factor", 2)
        return f"Zoomed in {img} at bbox_2d {bbox_2d} by {factor}x"
\end{lstlisting}

\paragraph{Crop.} This tool extracts a sub-region from the original image using specified 2D bounding box coordinates. It removes peripheral visual context to isolate and return the target area as a new image.

\begin{lstlisting}[language=Python]
class CropTool(BasicTool):
    name = "crop"
    description_en = "Crop a region from an image. Use localize_objects first if bbox is unknown."
    
    parameters = {
        "type": "object",
        "properties": {
            "image": {
                "type": "string",
                "description": "Image ID (e.g., 'img_0')"
            },
            "bbox_2d": {
                "type": "array",
                "items": {"type": "number"},
                "minItems": 4,
                "maxItems": 4,
                "description": "Bounding box as [left, top, right, bottom], values between 0 and 1"
            }
        },
        "required": ["image", "bbox_2d"]
    }
    example = '{"image": "img_0", "bbox_2d": [0.1, 0.2, 0.5, 0.6]}'
    
    def call(self, params: Union[str, Dict]) -> Dict:
        params_dict = self.parse_params(params)
        
        image_path = params_dict["image"]
        bbox_2d = params_dict["bbox_2d"]
        
        if not all(0 <= x <= 1.0 for x in bbox_2d):
            return {"error": "bbox_2d values must be between 0 and 1"}
        if bbox_2d[0] >= bbox_2d[2] or bbox_2d[1] >= bbox_2d[3]:
            return {"error": "invalid bbox_2d: left >= right or top >= bottom"}
        
        try:
            image = image_processing(image_path)
        except Exception as e:
            return {"error": f"failed to load image: {e}"}
        
        W, H = image.size
        crop_box = (int(bbox_2d[0] * W), int(bbox_2d[1] * H), int(bbox_2d[2] * W), int(bbox_2d[3] * H))
        cropped = image.crop(crop_box)
        
        return {"output_image": cropped}
    
    def generate_description(self, properties, observation):
        img = properties.get("image", "image")
        bbox_2d = properties.get("bbox_2d", [])
        return f"Cropped {img} at bbox_2d {bbox_2d}"
\end{lstlisting}

\paragraph{Visualize Regions.} This tool overlays annotated bounding boxes and their corresponding text labels directly onto an input image. It generates a new visual output that explicitly highlights the specified regions of interest for structural grounding.

\begin{lstlisting}[language=Python]
class VisualizeRegionsOnImageTool(BasicTool):
    name = "visualize_regions"
    description_en = "Draw bounding boxes and labels on an image. Each region is defined by normalized coordinates [x1, y1, x2, y2] where values are between 0 and 1. Use localize_objects first if bbox is unknown."
    
    parameters = {
        "type": "object",
        "properties": {
            "image": {
                "type": "string",
                "description": "Image ID (e.g., 'img_0')"
            },
            "regions": {
                "type": "array",
                "description": "List of regions to label, each with 'bbox_2d' and optional 'label'",
                "items": {
                    "type": "object",
                    "properties": {
                        "bbox_2d": {
                            "type": "array",
                            "description": "Normalized bounding box [x1, y1, x2, y2] where values are between 0 and 1",
                            "items": {"type": "number"},
                            "minItems": 4,
                            "maxItems": 4
                        },
                        "label": {
                            "type": "string",
                            "description": "Optional label text for the region"
                        }
                    },
                    "required": ["bbox_2d"]
                }
            },
            "color": {
                "type": "string",
                "description": "Color of the bounding boxes (default: 'yellow')"
            },
            "width": {
                "type": "integer",
                "description": "Width of the bounding box lines (default: 4)"
            }
        },
        "required": ["image", "regions"]
    }
    example = '{"image": "img_0", "regions": [{"bbox_2d": [0.1, 0.1, 0.5, 0.5], "label": "Object A"}, {"bbox_2d": [0.6, 0.6, 0.9, 0.9], "label": "Object B"}]}'
    
    def call(self, params: Union[str, Dict]) -> str:
        if not PIL_AVAILABLE:
            return {"error": "PIL (Pillow) is not installed. Please install it with: pip install Pillow"}
        
        params_dict = self.parse_params(params)
        
        image_path = params_dict["image"]
        regions = params_dict["regions"]
        color = params_dict.get("color", "yellow")
        width = params_dict.get("width", 4)
        
        try:
            image = image_processing(image_path)
            W, H = image.size
            
            img_labeled = image.copy()
            draw = ImageDraw.Draw(img_labeled)
            
            try:
                font = ImageFont.truetype('/usr/share/fonts/truetype/dejavu/DejaVuSansMono-Bold.ttf', 16)
            except (OSError, IOError):
                try:
                    font = ImageFont.truetype('/System/Library/Fonts/Helvetica.ttc', 16)
                except (OSError, IOError):
                    font = ImageFont.load_default()
            
            text_color = 'black'
            
            for obj in regions:
                bbox_2d = obj['bbox_2d']
                bbox_pixels = (bbox_2d[0] * W, bbox_2d[1] * H, bbox_2d[2] * W, bbox_2d[3] * H)
                
                draw.rectangle(bbox_pixels, outline=color, width=width)
                
                x1, y1, x2, y2 = bbox_pixels
                label = obj.get('label', '')
                
                if label:
                    try:
                        bbox_text = draw.textbbox((0, 0), label, font=font)
                        w = bbox_text[2] - bbox_text[0]
                        h = bbox_text[3] - bbox_text[1]
                    except AttributeError:
                        w, h = font.getsize(label)
                    
                    if x1 + w > W or y2 + h > H:
                        draw.rectangle((x1, y2 - h, x1 + w, y2), fill=color)
                        draw.text((x1, y2 - h), label, fill=text_color, font=font)
                    else:
                        draw.rectangle((x1, y2, x1 + w, y2 + h), fill=color)
                        draw.text((x1, y2), label, fill=text_color, font=font)
            
            return {"output_image": img_labeled}
            
        except FileNotFoundError:
            return {"error": f"Image file not found at {image_path}"}
        except Exception as e:
            return {"error": str(e)}
    
    def generate_description(self, properties, observation):
        img = properties.get("image", "image")
        regions = properties.get("regions", [])
        num_regions = len(regions) if isinstance(regions, list) else 0
        return f"Visualized {num_regions} regions on {img}"
\end{lstlisting}

\subsection{Visual Perception Tools}

\paragraph{OCR.} Optical Character Recognition scans an input image to detect and extract visible text elements. It returns the recognized textual content to assist in reading and parsing tasks.

\begin{lstlisting}[language=Python]
OCR_PROMPT = "OCR:"
MAX_PIXELS_OCR = 1280 * 28 * 28

class OCRTool(ModelBasedTool):
    name = "ocr"
    model_id = "ocr"

    description_en = "Extract text from an image. Returns empty if no text found."

    parameters = {
        "type": "object",
        "properties": {
            "image": {
                "type": "string",
                "description": "Image ID (e.g., 'img_0')"
            }
        },
        "required": ["image"]
    }
    example = '{"image": "img_0"}'

    def load_model(self, device: str) -> None:
        self.model = AutoModelForImageTextToText.from_pretrained(
            PADDLE_OCR_VL_MODEL_PATH, torch_dtype=torch.bfloat16
        ).to(device).eval()
        self.processor = AutoProcessor.from_pretrained(PADDLE_OCR_VL_MODEL_PATH)
        self.device = device
        self.is_loaded = True

    def _call_impl(self, params: Union[str, Dict]) -> Dict:
        params_dict = self.parse_params(params)
        image_path = params_dict["image"]
        tmp_file_to_remove = None

        try:
            full_path = image_path if os.path.exists(image_path) else get_full_path_data(image_path)
            if full_path and os.path.exists(full_path):
                image_file = full_path
            else:
                image = image_processing(image_path)
                if isinstance(image, Image.Image):
                    tmp = tempfile.NamedTemporaryFile(suffix=".png", delete=False)
                    tmp_file_to_remove = tmp.name
                    try:
                        image.save(tmp.name)
                        image_file = tmp.name
                    finally:
                        tmp.close()
                elif isinstance(image, str):
                    resolved = image if os.path.exists(image) else get_full_path_data(image)
                    if resolved is None:
                        raise FileNotFoundError(f"Image not found: {image_path}")
                    image_file = resolved
                else:
                    raise ValueError(f"Unexpected image type: {type(image)}")

            image = Image.open(image_file).convert("RGB")

            messages = [
                {
                    "role": "user",
                    "content": [
                        {"type": "image", "image": image},
                        {"type": "text", "text": OCR_PROMPT},
                    ]
                }
            ]
            inputs = self.processor.apply_chat_template(
                messages,
                add_generation_prompt=True,
                tokenize=True,
                return_dict=True,
                return_tensors="pt",
                images_kwargs={
                    "size": {
                        "shortest_edge": self.processor.image_processor.min_pixels,
                        "longest_edge": MAX_PIXELS_OCR,
                    }
                },
            )

            if "pixel_values" in inputs and not isinstance(inputs["pixel_values"], torch.Tensor):
                inputs["pixel_values"] = torch.from_numpy(inputs["pixel_values"]).to(self.device)
            if "image_grid_thw" in inputs and not isinstance(inputs["image_grid_thw"], torch.Tensor):
                inputs["image_grid_thw"] = torch.from_numpy(inputs["image_grid_thw"]).to(self.device)

            inputs = {
                k: v.to(self.device) if isinstance(v, torch.Tensor) else v
                for k, v in inputs.items()
            }

            outputs = self.model.generate(**inputs, max_new_tokens=512)
            extracted_text = self.processor.decode(
                outputs[0][inputs["input_ids"].shape[-1] : -1]
            ).strip()

            return {"extracted text": extracted_text}
        except FileNotFoundError as e:
            return {"error": f"Image not found: {str(e)}"}
        except Exception as e:
            return {"error": f"OCR error: {str(e)}\n{traceback.format_exc()}"}
        finally:
            if tmp_file_to_remove and os.path.exists(tmp_file_to_remove):
                try:
                    os.unlink(tmp_file_to_remove)
                except OSError:
                    pass
\end{lstlisting}

\paragraph{Localize Objects.} This tool identifies the spatial locations of user-specified object categories within an image. It processes the semantic queries and outputs the 2D bounding boxes corresponding to the detected entities.
\begin{lstlisting}[language=Python]
class LocalizeObjectsTool(ModelBasedTool):
    name = "localize_objects"
    model_id = "sam3"
    
    description_en = "Localize objects and return their bounding boxes. Use this to get bbox for other region-based tools. Note: Cannot detect text labels or annotation markers (e.g., 'A', 'B', 'point 1') drawn on images."
    
    parameters = {
        "type": "object",
        "properties": {
            "image": {"type": "string", "description": "Image ID (e.g., 'img_0')"},
            "objects": {"type": "array", "items": {"type": "string"}, "description": "A list of object names to localize. e.g. ['dog', 'cat', 'person']."}
        },
        "required": ["image", "objects"]
    }
    example = '{"image": "img_0", "objects": ["dog", "cat"]}'

    def load_model(self, device: str) -> None:
        self.processor = Sam3Processor.from_pretrained(SAM3_MODEL_PATH)
        self.model = Sam3Model.from_pretrained(SAM3_MODEL_PATH).to(device)
        self.device = device
        self.is_loaded = True

    def _call_impl(self, params: Union[str, Dict]) -> str:
        params_dict = self.parse_params(params)
        image_path = params_dict["image"]
        objects = params_dict["objects"]
        
        if not isinstance(objects, list) or len(objects) == 0:
            return {
                "error": "objects must be a non-empty list of strings"
            }
            
        try:
            image = image_processing(image_path)
            W, H = image.size
            
            regions = []
            obj_cnt = {}
            
            for obj_name in objects:
                inputs = self.processor(images=image, text=obj_name, return_tensors="pt").to(self.device)
                
                with torch.no_grad():
                    outputs = self.model(**inputs)
                
                results = self.processor.post_process_instance_segmentation(
                    outputs,
                    threshold=0.5,
                    mask_threshold=0.5,
                    target_sizes=inputs.get("original_sizes").tolist()
                )[0]
                
                boxes = results["boxes"]
                scores = results["scores"]
                
                for box, score in zip(boxes, scores):
                    score_val = score.item()
                    if score_val < 0.50:
                        continue
                    
                    box_list = box.tolist()
                    bbox = [
                        round(float(box_list[0]) / W, 4),
                        round(float(box_list[1]) / H, 4),
                        round(float(box_list[2]) / W, 4),
                        round(float(box_list[3]) / H, 4)
                    ]
                    
                    obj_cnt[obj_name] = obj_cnt.get(obj_name, 0) + 1
                    label_out = f"{obj_name}-{obj_cnt[obj_name]}" if obj_cnt[obj_name] > 1 else obj_name
                    regions.append({
                        "label": label_out,
                        "bbox_2d": bbox,
                        "score": round(score_val, 4)
                    })
            
            visualize_tool = VisualizeRegionsOnImageTool()
            visualize_params = {
                "image": image_path,
                "regions": [{"bbox_2d": r["bbox_2d"], "label": r["label"]} for r in regions]
            }
            output_image_result = visualize_tool.call(visualize_params)
            output_image = output_image_result.get("output_image")
            
            return {
                "output_image": output_image,
                "regions": regions
            }
            
        except FileNotFoundError as e:
            return {"error": f"Image file not found: {str(e)}"}
        except Exception as e:
            return {"error": f"Error localizing objects: {str(e)}"}
    
    def generate_description(self, properties, observation):
        img = properties.get("image", "image")
        objects = properties.get("objects", [])
        if isinstance(objects, list):
            objects_str = ", ".join(objects) if objects else "objects"
        else:
            objects_str = str(objects)
        return f"Localized {objects_str} in {img}"
\end{lstlisting}

\paragraph{Estimate Region Depth.} This tool calculates the spatial depth values within a designated bounding box. It aggregates the pixel-level depth maps and outputs statistical results, such as the mean depth of the specified area.

\begin{lstlisting}[language=Python]
class EstimateRegionDepthTool(ModelBasedTool):
    name = "estimate_region_depth"
    model_id = "depth_anything"
    
    description_en = "Estimate depth of a region. Smaller value = closer. Use localize_objects first if bbox is unknown."
    
    parameters = {
        "type": "object",
        "properties": {
            "image": {
                "type": "string",
                "description": "Image ID (e.g., 'img_0')"
            },
            "bbox_2d": {
                "type": "array",
                "items": {"type": "number"},
                "minItems": 4,
                "maxItems": 4,
                "description": "Bounding box [x1, y1, x2, y2] in normalized coordinates (0-1)"
            },
            "mode": {
                "type": "string",
                "enum": ["mean", "min", "max"],
                "description": "Mode to compute depth: mean (average), min (closest), or max (farthest)",
                "default": "mean"
            }
        },
        "required": ["image", "bbox_2d"]
    }
    example = '{"image": "img_0", "bbox_2d": [0.1, 0.2, 0.5, 0.6], "mode": "mean"}'
    
    def load_model(self, device: str) -> None:
        model_id = DEPTH_ANYTHING_PATH
        self.image_processor = AutoImageProcessor.from_pretrained(model_id)
        self.model = AutoModelForDepthEstimation.from_pretrained(model_id)
        self.model.to(device)
        self.model.eval()
        self.device = device
        self.is_loaded = True
    
    def _call_impl(self, params: Union[str, Dict]) -> str:
        params_dict = self.parse_params(params)
        image_path = params_dict["image"]
        bbox_2d = params_dict["bbox_2d"]
        mode = params_dict.get("mode", "mean")
        
        if not all(0 <= x <= 1.0 for x in bbox_2d):
            return {"error": "bbox_2d values must be between 0 and 1"}
        if bbox_2d[0] >= bbox_2d[2] or bbox_2d[1] >= bbox_2d[3]:
            return {"error": "invalid bbox_2d: left >= right or top >= bottom"}
        
        try:
            image = image_processing(image_path)
            W, H = image.size

            x1 = int(bbox_2d[0] * W)
            y1 = int(bbox_2d[1] * H)
            x2 = int(bbox_2d[2] * W)
            y2 = int(bbox_2d[3] * H)

            inputs = self.image_processor(images=image, return_tensors="pt")
            inputs = {k: v.to(self.device) for k, v in inputs.items()}

            with torch.no_grad():
                outputs = self.model(**inputs)

            processed_outputs = self.image_processor.post_process_depth_estimation(
                outputs,
                target_sizes=[(H, W)],
            )
            depth_map = processed_outputs[0]["predicted_depth"].squeeze().detach().cpu().numpy()
            depth_map = depth_map.max() - depth_map

            region_depth = depth_map[y1:y2, x1:x2]
            
            if mode == "mean":
                depth_value = float(np.mean(region_depth))
            elif mode == "min":
                depth_value = float(np.min(region_depth))
            elif mode == "max":
                depth_value = float(np.max(region_depth))
            else:
                depth_value = float(np.mean(region_depth))
            
            return {"estimated depth": round(depth_value, 4)}
            
        except FileNotFoundError as e:
            return {"error": f"Image file not found: {str(e)}"}
        except Exception as e:
            return {"error": f"Error estimating depth: {str(e)}"}
\end{lstlisting}

\paragraph{Estimate Object Depth.} This tool predicts the depth value of a specific semantic object within the visual scene. It combines object recognition with monocular depth estimation to locate the target and determine its distance.

\begin{lstlisting}[language=Python]
class EstimateObjectDepthTool(BasicTool):
    name = "estimate_object_depth"
    description_en = "Estimate depth of an object by text description. Smaller value = closer."
    
    parameters = {
        "type": "object",
        "properties": {
            "image": {
                "type": "string",
                "description": "Image ID (e.g., 'img_0')"
            },
            "object": {
                "type": "string",
                "description": "A short description of the object to get the depth from"
            },
            "mode": {
                "type": "string",
                "enum": ["mean", "min", "max"],
                "description": "Mode to compute depth: mean (average), min (closest), or max (farthest)",
                "default": "mean"
            }
        },
        "required": ["image", "object"]
    }
    example = '{"image": "img_0", "object": "a black cat", "mode": "mean"}'
    
    def call(self, params: Union[str, Dict]) -> Dict:
        params_dict = self.parse_params(params)
        image_path = params_dict["image"]
        object_desc = params_dict["object"]
        mode = params_dict.get("mode", "mean")
        
        try:
            localize_tool_class = TOOL_REGISTRY.get("localize_objects")
            if localize_tool_class is None:
                return {
                    "error": "localize_objects tool is not available"
                }
            
            localize_tool = localize_tool_class()
            localize_params = {
                "image": image_path,
                "objects": [object_desc]
            }
            localize_result = localize_tool.call(localize_params)
            
            if isinstance(localize_result, str):
                localize_data = json.loads(localize_result)
            else:
                localize_data = localize_result
            
            if "error" in localize_data or len(localize_data.get("regions", [])) == 0:
                return {
                    "error": "Object not found"
                }
            
            regions = localize_data["regions"]
            best_match_idx = np.argmax([region.get("score", 0) for region in regions])
            bbox_2d = regions[best_match_idx]["bbox_2d"]
            
            estimate_depth_tool_class = TOOL_REGISTRY.get("estimate_region_depth")
            if estimate_depth_tool_class is None:
                return {"error": "estimate_region_depth tool is not available"}
            
            estimate_depth_tool = estimate_depth_tool_class()
            depth_params = {
                "image": image_path,
                "bbox_2d": bbox_2d,
                "mode": mode
            }
            return estimate_depth_tool.call(depth_params)
                
        except Exception as e:
            return {"error": f"Error estimating object depth: {str(e)}"}
\end{lstlisting}

\subsection{Semantic Similarity Tools}

\paragraph{Image-to-Image Similarity.} This tool computes the visual embedding distance between a source image and a set of reference images. It outputs similarity scores to evaluate the visual resemblance among different image inputs.

\begin{lstlisting}[language=Python]
class GetImageToImagesSimilarityTool(ModelBasedTool):
    name = "get_image2images_similarity"
    model_id = "clip"
    
    description_en = "Compute similarity between one image and multiple images."
    
    parameters = {
        "type": "object",
        "properties": {
            "image": {
                "type": "string",
                "description": "Image ID (e.g., 'img_0')"
            },
            "other_images": {
                "type": "array",
                "items": {"type": "string"},
                "description": "The other images to compare to the reference image"
            }
        },
        "required": ["image", "other_images"]
    }
    example = '{"image": "img_0", "other_images": ["img_1", "img_2"]}'
    
    def load_model(self, device: str) -> None:
        self.model, _, self.preprocess = open_clip.create_model_and_transforms(CLIP_VERSION, pretrained=CLIP_PRETRAINED)
        self.model.eval()
        self.model = self.model.to(device)
        self.tokenizer = open_clip.get_tokenizer(CLIP_VERSION)
        self.device = device
        self.is_loaded = True
    
    def _call_impl(self, params: Union[str, Dict]) -> str:
        params_dict = self.parse_params(params)
        image_path = params_dict["image"]
        other_images = params_dict["other_images"]
        
        if not isinstance(other_images, list) or len(other_images) == 0:
            return {
                "error": "other_images must be a non-empty list of image paths"
            }
        
        try:
            reference_image = image_processing(image_path)
            reference_image_tensor = self.preprocess(reference_image).unsqueeze(0).to(self.device)
            
            other_images_processed = []
            for img_path in other_images:
                img = image_processing(img_path)
                other_images_processed.append(img)
            
            other_images_tensor = torch.stack([
                self.preprocess(img).to(self.device) for img in other_images_processed
            ])
            
            with torch.no_grad(), torch.cuda.amp.autocast():
                reference_features = self.model.encode_image(reference_image_tensor)
                other_features = self.model.encode_image(other_images_tensor)
                
                reference_features /= reference_features.norm(dim=-1, keepdim=True)
                other_features /= other_features.norm(dim=-1, keepdim=True)
                
                similarity_scores = reference_features @ other_features.T
            
            sim_scores = [round(score.item(), 2) for score in similarity_scores[0]]
            
            best_image_index = torch.argmax(similarity_scores, dim=1).item()
            
            return {
                "similarity scores": sim_scores,
                "best match": other_images[best_image_index]
            }
            
        except FileNotFoundError as e:
            return {"error": f"Image file not found: {str(e)}"}
        except Exception as e:
            return {"error": f"Error computing image similarity: {str(e)}"}
\end{lstlisting}

\paragraph{Image-to-Text Similarity.} This tool evaluates the semantic alignment between a given image and multiple textual descriptions. It returns scores indicating how accurately each text matches the visual content.

\begin{lstlisting}[language=Python]
class GetImageToTextsSimilarityTool(ModelBasedTool):
    name = "get_image2texts_similarity"
    model_id = "clip"
    
    description_en = "Compute similarity between one image and multiple texts (image  texts)."
    
    parameters = {
        "type": "object",
        "properties": {
            "image": {
                "type": "string",
                "description": "Image ID (e.g., 'img_0')"
            },
            "texts": {
                "type": "array",
                "items": {"type": "string"},
                "description": "A list of texts to compare to the reference image"
            }
        },
        "required": ["image", "texts"]
    }
    example = '{"image": "img_0", "texts": ["a cat", "a dog"]}'
    
    def load_model(self, device: str) -> None:
        self.model, _, self.preprocess = open_clip.create_model_and_transforms(CLIP_VERSION, pretrained=CLIP_PRETRAINED)
        self.model.eval()
        self.model = self.model.to(device)
        self.tokenizer = open_clip.get_tokenizer(CLIP_VERSION)
        self.device = device
        self.is_loaded = True
    
    def _call_impl(self, params: Union[str, Dict]) -> str:
        params_dict = self.parse_params(params)
        image_path = params_dict["image"]
        texts = params_dict["texts"]
        
        if not isinstance(texts, list) or len(texts) == 0:
            return {
                "error": "texts must be a non-empty list of strings"
            }
        
        try:
            reference_image = image_processing(image_path)
            reference_image_tensor = self.preprocess(reference_image).unsqueeze(0).to(self.device)
            
            text_tokens = self.tokenizer(texts).to(self.device)
            
            with torch.no_grad(), torch.cuda.amp.autocast():
                image_features = self.model.encode_image(reference_image_tensor)
                text_features = self.model.encode_text(text_tokens)
                
                image_features /= image_features.norm(dim=-1, keepdim=True)
                text_features /= text_features.norm(dim=-1, keepdim=True)
                
                similarity_scores = image_features @ text_features.T
            
            sim_scores = [round(score.item(), 2) for score in similarity_scores[0]]
            
            best_text_index = torch.argmax(similarity_scores, dim=1).item()
            
            return {
                "similarity scores": sim_scores,
                "best match": texts[best_text_index]
            }
            
        except FileNotFoundError as e:
            return {"error": f"Image file not found: {str(e)}"}
        except Exception as e:
            return {"error": f"Error computing image-to-text similarity: {str(e)}"}
\end{lstlisting}

\paragraph{Text-to-Image Similarity.} This tool measures how well a set of images corresponds to a specific text query. It calculates alignment scores for each image, functioning as a targeted text-based image retrieval metric.

\begin{lstlisting}[language=Python]
class GetTextToImagesSimilarityTool(ModelBasedTool):
    name = "get_text2images_similarity"
    model_id = "clip"
    
    description_en = "Compute similarity between one text and multiple images (text  images)."
    
    parameters = {
        "type": "object",
        "properties": {
            "text": {
                "type": "string",
                "description": "The reference text"
            },
            "images": {
                "type": "array",
                "items": {"type": "string"},
                "description": "A list of images to compare to the reference text"
            }
        },
        "required": ["text", "images"]
    }
    example = '{"text": "a black and white cat", "images": ["img_0", "img_1"]}'
    
    def load_model(self, device: str) -> None:
        self.model, _, self.preprocess = open_clip.create_model_and_transforms(CLIP_VERSION, pretrained=CLIP_PRETRAINED)
        self.model.eval()
        self.model = self.model.to(device)
        self.tokenizer = open_clip.get_tokenizer(CLIP_VERSION)
        self.device = device
        self.is_loaded = True
    
    def _call_impl(self, params: Union[str, Dict]) -> str:
        params_dict = self.parse_params(params)
        text = params_dict["text"]
        images = params_dict["images"]
        
        if not isinstance(images, list) or len(images) == 0:
            return {
                "error": "images must be a non-empty list of image paths"
            }
        
        try:
            text_tokens = self.tokenizer([text]).to(self.device)
            
            images_processed = []
            for img_path in images:
                img = image_processing(img_path)
                images_processed.append(img)
            
            images_tensor = torch.stack([
                self.preprocess(img).to(self.device) for img in images_processed
            ])
            
            with torch.no_grad(), torch.cuda.amp.autocast():
                text_features = self.model.encode_text(text_tokens)
                image_features = self.model.encode_image(images_tensor)
                
                text_features /= text_features.norm(dim=-1, keepdim=True)
                image_features /= image_features.norm(dim=-1, keepdim=True)
                
                similarity_scores = text_features @ image_features.T
            
            sim_scores = [round(score.item(), 2) for score in similarity_scores[0]]
            
            best_image_index = torch.argmax(similarity_scores, dim=1).item()
            
            return {
                "similarity scores": sim_scores,
                "best match": images[best_image_index]
            }
            
        except FileNotFoundError as e:
            return {"error": f"Image file not found: {str(e)}"}
        except Exception as e:
            return {"error": f"Error computing text-to-image similarity: {str(e)}"}
\end{lstlisting}

\subsection{Reasoning and External Tools}

\paragraph{Calculator.} This tool evaluates standard arithmetic expressions provided as strings. It processes mathematical operators and numbers to return the exact numerical calculation result.

\begin{lstlisting}[language=Python]
class CalculatorTool(BasicTool):
    name = "calculator"
    description_en = "Calculate mathematical expressions. Supports basic arithmetic (+, -, *, /), exponentiation (**), and parentheses."
    
    parameters = {
        "type": "object",
        "properties": {
            "expression": {
                "type": "string",
                "description": "The mathematical expression to calculate"
            }
        },
        "required": ["expression"]
    }
    example = '{"expression": "123 * 456 + 789"}'
    
    def call(self, params: Union[str, Dict]) -> Dict:
        params_dict = self.parse_params(params)
        expression = params_dict["expression"]
        
        if not self._is_safe_expression(expression):
            return {"error": "Expression contains invalid or unsafe characters"}
        
        try:
            result = eval(expression, {"__builtins__": {}}, {})
            return {"calculation result": result}
        except ZeroDivisionError:
            return {"error": "Division by zero"}
        except SyntaxError:
            return {"error": "Invalid syntax in expression"}
        except Exception as e:
            return {"error": str(e)}
    
    def _is_safe_expression(self, expr: str) -> bool:
        safe_pattern = re.compile(r'^[\d\s\+\-\*\/\(\)\.\*\*]+$')
        return bool(safe_pattern.match(expr))
\end{lstlisting}

\paragraph{Solve Math Equation.} This tool processes and solves algebraic equations or math-related word problems presented in natural language formats. It executes the mathematical logic and outputs the analytical or numerical solution.

\begin{lstlisting}[language=Python]
class SolveMathEquationTool(BasicTool):
    name = "solve_math_equation"
    description_en = (
        "Solve mathematical equations and problems using WolframAlpha. "
        "Supports algebra, calculus, and symbolic/numeric math queries."
    )
    
    parameters = {
        "type": "object",
        "properties": {
            "query": {
                "type": "string",
                "description": "A mathematical question or equation to solve",
            }
        },
        "required": ["query"],
    }
    example = '{"query": "x^2 + 2x + 1 = 0, what is x?"}'

    _key_index = 0
    _key_lock = threading.Lock()

    @staticmethod
    def _load_wolfram_keys_from_env() -> List[str]:
        keys = []

        api_keys_str = os.getenv("WOLFRAM_ALPHA_API_KEYS", "").strip()
        if api_keys_str:
            for k in api_keys_str.split(","):
                k = k.strip()
                if k and k not in keys:
                    keys.append(k)
            return keys

        if not DOTENV_AVAILABLE:
            return keys

        project_root = Path(__file__).resolve().parent.parent
        candidate_env_files = [
            project_root / "api" / "utils" / "keys.env",
            project_root / "keys.env",
            project_root / ".env",
        ]
        for env_path in candidate_env_files:
            if env_path.exists():
                load_dotenv(env_path, override=False)
                api_keys_str = os.getenv("WOLFRAM_ALPHA_API_KEYS", "").strip()
                if api_keys_str:
                    for k in api_keys_str.split(","):
                        k = k.strip()
                        if k and k not in keys:
                            keys.append(k)
                if keys:
                    break

        return keys

    @classmethod
    def _get_next_key(cls, keys: List[str]) -> str:
        if not keys:
            return ""
        with cls._key_lock:
            key = keys[cls._key_index % len(keys)]
            cls._key_index = (cls._key_index + 1) % len(keys)
            return key

    def __init__(self, cfg=None, use_zh=False):
        super().__init__(cfg, use_zh)
        self.api_keys: List[str] = []
        self.max_retries = 3

        if cfg is not None:
            cfg_keys = cfg.get("wolfram_api_keys", [])
            if isinstance(cfg_keys, str):
                cfg_keys = [k.strip() for k in cfg_keys.split(",") if k.strip()]
            elif isinstance(cfg_keys, list):
                cfg_keys = [k.strip() for k in cfg_keys if isinstance(k, str) and k.strip()]
            self.api_keys = cfg_keys
            self.max_retries = cfg.get("wolfram_max_retries", 3)

        if not self.api_keys:
            self.api_keys = self._load_wolfram_keys_from_env()

    @staticmethod
    def _looks_like_interpretation(text: str) -> bool:
        if not text:
            return True
        t = text.strip().lower()
        patterns = (
            "input interpretation",
            "solve ",
            " for x",
            " for y",
            " for z",
        )
        return any(p in t for p in patterns)

    @staticmethod
    def _parse_xml_response(xml_text: str) -> Dict:
        root = ET.fromstring(xml_text)
        success_raw = root.attrib.get("success", "false")
        success = str(success_raw).lower() == "true"
        if not success:
            return {"error": "Your Wolfram query is invalid. Please try a new query."}

        answer = ""
        for pod in root.findall("pod"):
            title = pod.attrib.get("title", "")
            if title == "Solution":
                subpods = pod.findall("subpod")
                if subpods:
                    plaintext = subpods[0].findtext("plaintext") or ""
                    answer = plaintext
            if title in {"Results", "Solutions"}:
                subpods = pod.findall("subpod")
                for i, sub in enumerate(subpods):
                    text = sub.findtext("plaintext") or ""
                    answer += f"ans {i}: {text}\n"
                break

        if not answer:
            preferred_titles = {
                "Result",
                "Results",
                "Solution",
                "Solutions",
                "Exact result",
                "Decimal approximation",
            }
            for pod in root.findall("pod"):
                title = pod.attrib.get("title", "")
                if title not in preferred_titles:
                    continue
                sub = pod.find("subpod")
                if sub is not None:
                    text = (sub.findtext("plaintext") or "").strip()
                    if text and not SolveMathEquationTool._looks_like_interpretation(text):
                        answer = text
                        break

        if not answer or SolveMathEquationTool._looks_like_interpretation(answer):
            return {"error": "No good Wolfram Alpha result was found."}
        return {"result": answer.strip()}

    @staticmethod
    def _is_retryable_error(error_msg: str) -> bool:
        error_lower = error_msg.lower()
        retryable_patterns = ["rate limit", "quota", "too many requests", "403", "429"]
        return any(p in error_lower for p in retryable_patterns)

    def _query_via_http(self, query: str, api_key: str) -> Dict:
        try:
            resp = requests.get(
                "https://api.wolframalpha.com/v2/query",
                params={"appid": api_key, "input": query},
                timeout=30,
            )
            if resp.status_code != 200:
                return {
                    "error": (
                        f"WolframAlpha HTTP error: code={resp.status_code}, "
                        f"body_preview={resp.text[:240]}"
                    ),
                    "_retryable": resp.status_code in (403, 429),
                }
            return self._parse_xml_response(resp.text)
        except Exception as e:
            return {
                "error": f"HTTP request failed: {type(e).__name__}: {e}",
                "_retryable": self._is_retryable_error(str(e)),
            }

    def _query_with_retry(self, query: str) -> Dict:
        tried_keys = set()
        last_error = None

        for _ in range(min(self.max_retries, len(self.api_keys))):
            current_key = self._get_next_key(self.api_keys)
            if current_key in tried_keys:
                continue
            tried_keys.add(current_key)

            result = self._query_via_http(query, current_key)
            if "error" not in result or not result.pop("_retryable", False):
                return result
            last_error = result

        if last_error:
            last_error.pop("_retryable", None)
        return last_error or {"error": "All API keys exhausted."}

    def call(self, params: Union[str, Dict]) -> Dict:
        params_dict = self.parse_params(params)
        query = params_dict["query"].strip()

        if not query:
            return {"error": "Query cannot be empty."}

        if not self.api_keys:
            return {
                "error": (
                    "WOLFRAM_ALPHA_API_KEYS is not set. "
                    "Set it in config or environment (.env / keys.env)."
                )
            }

        return self._query_with_retry(query)
\end{lstlisting}

\paragraph{Web Search.} This tool queries external search engines using provided text keywords to retrieve up-to-date factual information. It returns the top relevant text snippets to supplement the internal knowledge of the agent.

\begin{lstlisting}[language=Python]
class WebSearchTool(BasicTool):
    name = "web_search"
    description_en = (
        "Search the web for information using a text query. "
        "Returns the top-k results, each with a title and snippet."
    )
    
    parameters = {
        "type": "object",
        "properties": {
            "query": {
                "type": "string",
                "description": "Search query text",
            },
            "top_k": {
                "type": "integer",
                "description": "Number of results to return (1-5, default: 3)",
            },
        },
        "required": ["query"],
    }
    example = '{"query": "Which city is the capital of China?", "top_k": 3}'

    def __init__(self, cfg: Optional[Dict] = None, use_zh: bool = False):
        super().__init__(cfg, use_zh)
        self._max_retries = 3
        self._retry_delay = 2

    def call(self, params: Union[str, Dict]) -> Dict:
        params_dict = self.parse_params(params)
        query = params_dict.get("query", "").strip()
        top_k = max(1, min(int(params_dict.get("top_k", 3)), 5))

        if not query:
            return {"error": "Parameter 'query' is required"}

        try:
            return self._text_search(query, top_k)
        except Exception as e:
            logging.error(f"Web search failed: {e}")
            return {"error": f"Search failed: {str(e)}"}

    async def call_async(self, params: Union[str, Dict]) -> Dict:
        return await asyncio.to_thread(self.call, params)

    def _text_search(self, query: str, top_k: int) -> Dict:
        url = "https://google.serper.dev/search"
        headers = {
            "X-API-KEY": GOOGLE_SEARCH_API_KEY,
            "Content-Type": "application/json",
        }
        payload = {
            "q": query,
            "gl": "cn",
            "hl": "zh-cn",
            "location": "China",
            "num": top_k,
        }

        for attempt in range(self._max_retries):
            try:
                resp = requests.post(
                    url, headers=headers, data=json.dumps(payload), timeout=30
                )
                resp.raise_for_status()
                data = resp.json()
                results = data.get("organic", [])[:top_k]

                if not results:
                    return {"search results": "No results found"}

                lines = []
                for i, res in enumerate(results, 1):
                    title = res.get("title", "")
                    snippet = res.get("snippet", "")
                    lines.append(f"[{i}] Title: {title}")
                    lines.append(f"Snippet: {snippet}")
                    lines.append("")

                return {"search results": "\n".join(lines).strip()}
            except Exception as e:
                logging.warning(
                    f"Request failed (attempt {attempt + 1}/{self._max_retries}): {e}"
                )
                if attempt < self._max_retries - 1:
                    time.sleep(self._retry_delay)
                else:
                    raise
\end{lstlisting}

\section{Prompts}
\label{app:prompt}

\subsection{Evaluation Prompt}

\begin{promptbox}{Evaluation Prompt}
\label{box:multimodaltoolassistant}

\textbf{Role:} You are a multimodal assistant with tool access and memory.

\medskip

\textbf{Task:} Analyze images and questions carefully, decide whether tool usage is necessary, and use tools strategically when they help obtain information that cannot be reliably determined through direct observation.

\medskip

\textbf{Tool Usage Guidelines:}

Before using tools, analyze and plan:

\begin{itemize}[leftmargin=*, itemsep=0.2em]
    \item Carefully analyze the task requirements and image content
    \item Identify what information is required to answer the question
    \item Evaluate which tools are suitable for the current task
    \item Plan the sequence of tool usage if multiple tools are needed
    \item Consider how outputs from multiple tools can be combined to derive the final answer
\end{itemize}

\textbf{When tools are appropriate:}

\begin{itemize}[leftmargin=*, itemsep=0.2em]
    \item When examining details that are difficult to see directly (e.g., small text, tiny objects, subtle textures)
    \item When precise localization or counting is required
    \item When the image content is complex and requires step-by-step regional analysis
\end{itemize}

\textbf{When tools are not appropriate:}

\begin{itemize}[leftmargin=*, itemsep=0.2em]
    \item When cropping or zooming would remove important contextual information
    \item When the answer can already be determined through direct observation
\end{itemize}

\textbf{Best practices when using tools:}

\begin{itemize}[leftmargin=*, itemsep=0.2em]
    \item Clear objectives: know exactly what information you want before calling a tool
    \item Accurate regions: ensure the selected region corresponds to the part that requires analysis
    \item Analyze outputs critically and combine them with direct observation
    \item Do not replace visible evidence with speculative inference
\end{itemize}

If a tool call fails or returns an error, adapt by trying alternative tools or adjusting your strategy.

\medskip

\textbf{Image Reference:}

All images are referenced with unique IDs (img\_0, img\_1, ...).  
When calling tools, use these IDs to specify which image to operate on.

\medskip

\textbf{[If stateful experience search is enabled]:}

\textbf{Experience Retrieval:}

Before each action step, you MUST call \texttt{search\_experiences} to retrieve advice from similar previously solved tasks.

\begin{itemize}[leftmargin=*, itemsep=0.2em]
    \item Choose a retrieval view that best matches the current information need
    \item Multiple retrieval calls are allowed if necessary
    \item Use retrieved advice to guide decision making before taking the next action
\end{itemize}

\end{promptbox}

\subsection{Prompt for Stateful Experience Abstraction}

\begin{promptbox}{Prompt for Stateful Experience Abstraction}
\label{box:hindsight_reasoning}

\textbf{Role:} You are evaluating a complete agent reasoning trace.

\textbf{[If the task was solved CORRECTLY]:}
The task was solved CORRECTLY. The process may be highly efficient, or it may contain wasteful, redundant, or error-prone steps. Judge each step by the quality and relevance of its output.

\textbf{[If the task was solved INCORRECTLY]:}
The task was solved INCORRECTLY. Correct answer: \{ground\_truth\}.
Your goal is to identify which steps were most responsible for the incorrect outcome. A wrong final answer does NOT mean every step was bad — many steps may have been perfectly reasonable. Focus on pinpointing the critical steps that introduced errors, ignored evidence, or led the reasoning astray.

\medskip

\textbf{Task:}

\{images\_note\} \\
\textbf{Question:} \{question\} \{type\_line\} \\
\{tools\_section\}

\medskip

\textbf{Trajectory:}

The initial state $s_0$ is the question and images above. Each subsequent state is what the agent has seen so far, including all previous actions and observations.

\{traj\_text\}

\medskip

\textbf{Instructions:}

\textbf{[If the task was solved CORRECTLY]:}
For each (state, action) pair, provide:

\textbf{1. q\_value (0-10):} 
Rate the quality of this action at this state. Judge whether it produced accurate, relevant information that actually contributed to reaching the answer.
\begin{itemize}[leftmargin=*, itemsep=0.2em]
    \item \textbf{9-10 (Essential):} Produced decisive information that directly shaped the answer, or answered correctly at the right time
    \item \textbf{7-8 (Helpful):} Useful output that advanced the solution, with minor room for improvement in tool choice or parameters
    \item \textbf{5-6 (Reasonable):} Valid approach, but output had little actual influence on the final answer; a more direct path existed
    \item \textbf{3-4 (Wasteful):} Produced no new information, duplicated prior knowledge, or was an unnecessary detour
    \item \textbf{0-2 (Harmful):} Produced errors, misleading output, or repeated a known failure
\end{itemize}

\textbf{[If the task was solved INCORRECTLY]:}
Rate how much this step CONTRIBUTED TO the incorrect outcome. A high score means this step was a critical cause of the error.
\begin{itemize}[leftmargin=*, itemsep=0.2em]
    \item \textbf{9-10 (Critical error):} Directly caused or cemented the wrong answer (e.g., misread evidence, wrong tool, flawed conclusion)
    \item \textbf{7-8 (Significant):} Substantially misled the reasoning, even if not the sole cause of failure
    \item \textbf{5-6 (Moderate):} Somewhat contributed to the error, but other steps were more decisive
    \item \textbf{3-4 (Minor):} Had little impact on the incorrect outcome; the step was mostly neutral
    \item \textbf{0-2 (Irrelevant):} This step was reasonable and did not contribute to the failure at all
\end{itemize}

\textbf{2. experience (1-2 sentences):} Actionable advice that a FUTURE agent would see BEFORE making this decision. This advice will be injected into the agent's context at runtime to guide its next action.

\textbf{Guidelines:}
\begin{itemize}[leftmargin=*, itemsep=0.2em]
    \item Reference the task type and tool strategy, but keep the advice generalizable — do not mention specific objects, labels, or values from this particular trace.
    \item NEVER mention or hint at the correct answer.
    \item \textbf{[If the task was solved CORRECTLY]:} Focus on STRATEGY: recommend the specific tool and approach that works, or advise answering directly if tools are unnecessary. If the step was an error, redundant, or unnecessary, the experience MUST be cautionary.
    \item \textbf{[If the task was solved INCORRECTLY]:} For high \texttt{q\_value} steps (critical errors), the experience MUST be a cautionary warning — explain what went wrong and what to avoid. For low \texttt{q\_value} steps, provide brief neutral or positive guidance.
\end{itemize}

\textbf{Examples:}
\begin{itemize}[leftmargin=*, itemsep=0.2em]
    \item \textbf{Good (Strategy):} "For visual similarity tasks, use \texttt{get\_image2images\_similarity} to get objective scores rather than estimating visually."
    \item \textbf{Good (Direct Answer):} "The observation already contains enough information to answer directly. No further tool calls are needed."
    \item \textbf{Good (Correction):} "When observations conflict with prior reasoning, re-examine the evidence before committing to an answer."
    \item \textbf{Bad (Too Vague):} "The agent should use the right tool." / "The agent should be more careful."
    \item \textbf{Bad (Leaks Answer):} "The answer is likely A based on the scores."
\end{itemize}

\medskip

\textbf{Output Format:}

Output JSON array: \\
\texttt{[\{"state": 0, "q\_value": 8, "experience": "..."\}, ...]}

\end{promptbox}

\subsection{Prompt for Trajectory-level Experience Abstraction}

\begin{promptbox}{Prompt for Trajectory-level Experience Abstraction}
\label{box:trajectory_experience}

\textbf{Role:} You are evaluating a complete agent reasoning trace.

\medskip

\textbf{[If the task was solved CORRECTLY]:}
The task was solved CORRECTLY. Your goal is to summarize what strategy worked well across the reasoning trajectory. Focus on effective tool usage, evidence interpretation, and reasoning strategies that helped reach the correct answer.

\textbf{[If the task was solved INCORRECTLY]:}
The task was solved INCORRECTLY. Your goal is to identify what went wrong in the reasoning trajectory and what a future agent should do differently to avoid similar mistakes.

\medskip

\textbf{Task:}

\{images\_note\} \\
\textbf{Question:} \{question\} \{type\_line\} \\
\{tools\_section\}

\medskip

\textbf{Trajectory:}

\{traj\_text\}

\medskip

\textbf{Instructions:}

Write exactly \textbf{1--2 concise sentences} of actionable experience for future agents facing similar tasks.

\textbf{[If the task was solved CORRECTLY]:}

\begin{itemize}[leftmargin=*, itemsep=0.2em]
    \item Summarize the strategy that worked well (e.g., tool choice, evidence usage, reasoning approach)
    \item Highlight patterns that helped reach the correct answer efficiently
    \item Focus on practices that future agents should replicate
\end{itemize}

\textbf{[If the task was solved INCORRECTLY]:}

\begin{itemize}[leftmargin=*, itemsep=0.2em]
    \item Identify what went wrong in tool usage, reasoning, or evidence interpretation
    \item Explain what a future agent should do differently
    \item Highlight common mistakes or misleading reasoning patterns to avoid
\end{itemize}

\textbf{General Guidelines:}

\begin{itemize}[leftmargin=*, itemsep=0.2em]
    \item Keep the advice generalizable to similar tasks
    \item Do not mention specific objects, labels, or values from the trajectory
    \item NEVER reveal or hint at the correct answer
\end{itemize}

\medskip

\textbf{Output Format:}

Output JSON object: \\
\texttt{\{"experience": "your 1-2 sentence experience"\}}

\end{promptbox}

\end{document}